\documentclass[runningheads]{llncs}
\usepackage{graphicx}
\usepackage{amsmath,amssymb} 
\usepackage{color}

\usepackage{makeidx} 
\usepackage{subfiles} 
\usepackage{tabu} 
\usepackage{array,multirow}
\usepackage{subcaption} 
\usepackage[utf8x]{inputenc} 
 
\usepackage{xfrac}
\usepackage{floatrow} 
\newcommand{\etal}{\textit{et al}.}
\newcommand{\ie}{\textit{i}.\textit{e}.}
\newcommand{\eg}{\textit{e}.\textit{g}.}
\newcommand{\etc}{\textit{etc}.}

\begin{document}
\title{Learning Style Compatibility for Furniture} 

\titlerunning{Learning Style Compatibility for Furniture}

\author{Divyansh Aggarwal\inst{1} \and
Elchin Valiyev\inst{2} \and
Fadime Sener\inst{2} \and
Angela Yao\inst{2}  
}
%
\authorrunning{D. Aggarwal, E. Valiyev, F. Sener, A. Yao}

\institute{IIT Jodhpur, India \and
University of Bonn, Germany\\ 
\email{aggarwal.1@iitj.ac.in}\\ 
\email{s6elvali@uni-bonn.de, \{sener,yao\}@cs.uni-bonn.de}}
\maketitle              
\begin{abstract}
When judging style, a key question that often arises is whether or not a pair of objects are compatible with each other. In this paper we investigate how Siamese networks can be used efficiently for assessing the style compatibility between images of furniture items. We show that the middle layers of pretrained CNNs can capture essential information about furniture style, which allows for efficient applications of such networks for this task. We also use a joint image-text embedding method that allows for the querying of stylistically compatible furniture items, along with additional attribute constraints based on text. To evaluate our methods, we collect and present a large scale dataset of images of furniture of different style categories accompanied by text attributes.

\end{abstract}

\section{Introduction}
\label{sec:introduction}

Understanding visual styles is important for application domains such as art, cinematography, advertising, and so on. Previous research in this domain focused primarily on recognition, \eg~of photo aesthetic quality~\cite{datta}, building styles~\cite{architecture}, illustration and painting styles~\cite{illustration_style,art,painter}, city architectures~\cite{doersch2012makes}, clothing styles ~\cite{kiapour2014hipster}. Beyond recognition, however, style becomes a difficult concept to quantify, since it is a subjective and fine-grained problem. In this paper, we address the problem of style compatibility. We ask whether two objects -- furniture specifically, are stylistically congruent, \eg\emph{``How well does this chair match that table?"}. Answering such a question would be greatly useful for supporting interior design, but is highly challenging, as it requires the ability to recognize features and attributes characteristic of specific styles across multiple object categories. 

Recently, a few works have addressed style compatibility, though their focus is on clothing~\cite{mcauley2015image,veit2015learning,bilstm}. Our work is inspired by~\cite{veit2015learning}, which uses a Siamese network to measure compatibility between different types of clothing. Siamese networks have shown great success in measuring fine-grained visual similarities~\cite{wang2014learning,lin2015learning,similarity}. In this work, we investigate the use of different Siamese architectures in assessing furniture style compatibility.

Style, as a high-level semantic concept, can be conveyed not only through the visual appearance, but also through text descriptions. As style data is often compiled from retail resources on the web~\cite{similarity,bilstm}, the text tags and meta-data that often accompany images can also serve as an additional source of information. As such, we also use a joint visual-text embedding for learning stylistic compatibility. Such a model can make stylistically compatible recommendations in the presence of constraints such as the type of furniture, material and color.

To evaluate our methods, we collect and present the Bonn Furniture Styles Dataset. This is a large-scale dataset of approximately 90,000 images of furniture, drawn from six types and 17 different styles. To the best of our knowledge, the only dataset of furniture to date is the Singapore Furniture Dataset from Hu~\etal~\cite{style_classification} and is too small ($\sim$3000 images) to effectively learn deep models. Our dataset is more challenging and in addition, each image is accompanied by text descriptions of the furniture, with key attributes such as material, color, and so on.  

\begin{figure}[t!]
\centering
\includegraphics[width = 0.9\linewidth]{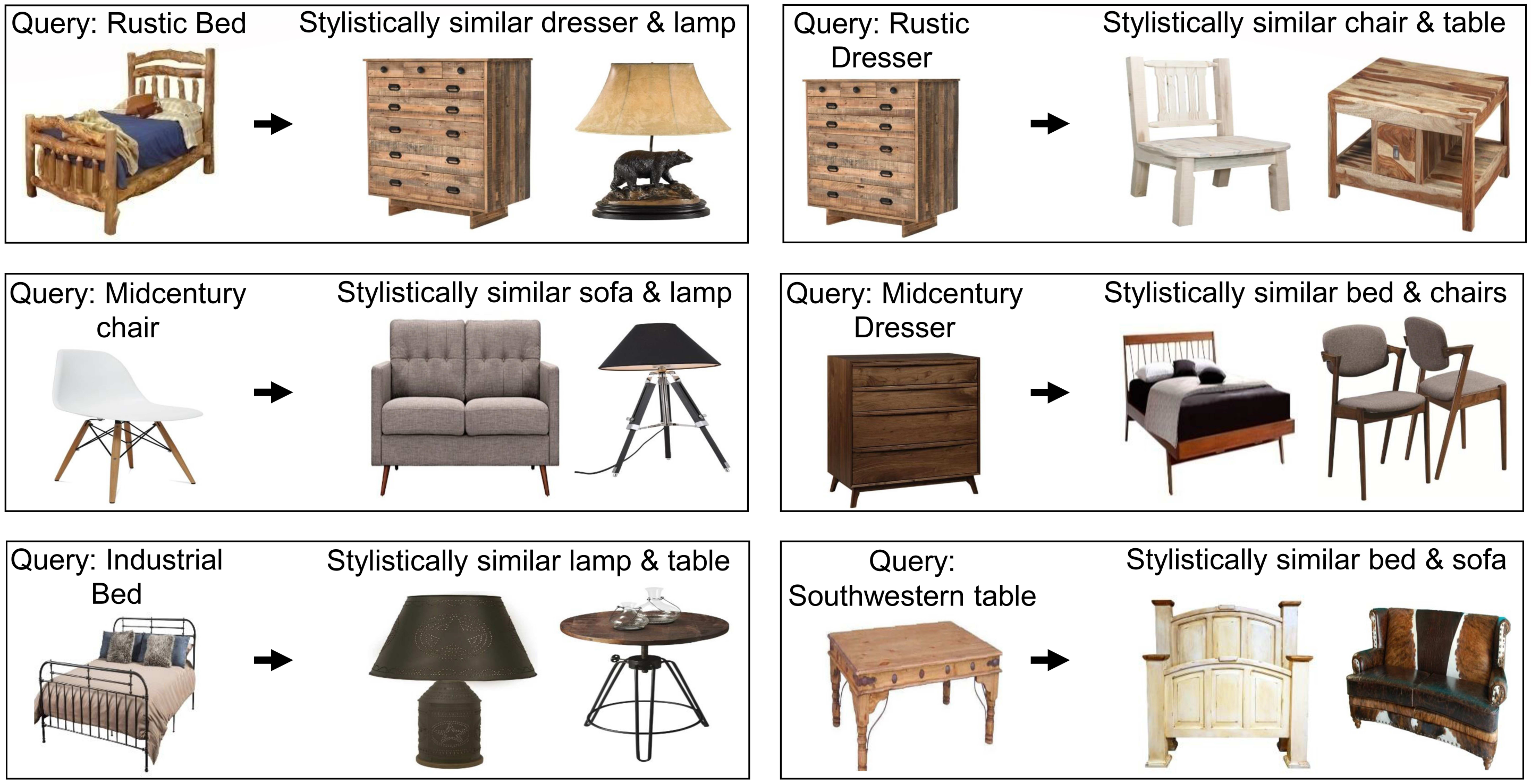}
\caption{Applying our categorical Siamese model (see Sec.~\ref{sec:catsiam}), we show the two most stylistically compatible furniture items with respect to the query item. Note that our model is specifically trained across different types of furniture.}
\label{siamese_compatibility_search}
\end{figure}

We experiment using the Singapore Furniture Dataset~\cite{style_classification} as well as our own large-scale dataset. Our findings show that Siamese networks are not only very successful at evaluating furniture style compatibility but also that compatibility can  be assessed with only mid-level features. This opens up the possibility of deploying such a network efficiently on resource-constrained platforms and allows individuals to assess furniture compatibility in their own home with mobile devices. Our main contributions can be summarized as follows:

\begin{itemize}
\item We explore Siamese networks for learning the style compatibility of furniture and show that style, as a high-level semantic concept, is sufficiently encoded by the middle layers of convolutional networks. As such, style compatibility can already be assessed using relatively short CNNs without much loss in performance in comparison to full-depth CNNs. 
\item We use a joint visual-text embedding which can make style-aware furniture recommendations with additional text-based constraints.
\item We collected a new large-scale dataset of furniture of different styles, along with associated text descriptions such as category, color, material etc. The dataset will be made publicly available and will be of interest for those working not only on style understanding but also on image attributes and fine-grained image understanding.
\end{itemize}
 
\section{Related Work} 
\label{chapter_relatedwork}

Recognizing visual styles in computer vision has been addressed for classifying classic painting styles~\cite{art,painter}, building architectures~\cite{architecture} and illustrations and photos~\cite{illustration_style,karayev2013recognizing}. Style has also been explored within the context of predicting the aesthetic quality of photographs~\cite{datta} and paintings~\cite{aesthetic}, image memorability~\cite{memorable} and image ``interestingness''~\cite{interestingness}. 
Exploring stylistic similarity on clothing ~\cite{kiapour2014hipster,neuroaesthetics,wherebuy,veit2015learning,fashion_128} has drawn a lot attention due to the broad potential for commercial applications. For example, Veit~\etal~\cite{veit2015learning} showed that Siamese networks can assess clothing style compatibility. Likewise, in~\cite{bilstm}, a bidirectional LSTM and Inception-V3~\cite{szegedy2016rethinking} model was used for visual-semantic embedding of outfit images and their descriptions. The learned embedding are then used for outfit generation, compatibility prediction and finding matching item tasks. Note however, such a method requires sets or ensembles of fashion items for training compatibility. 

Within the realm of furniture and home decoration a few works on style have been proposed~\cite{similarity,style_classification,furniture_style}. Bell~\etal\cite{similarity} evaluated Siamese networks for learning visual similarity between iconic images of furniture depicted on white backgrounds versus in actual scenes. More recently, Hu~\etal~\cite{style_classification} presented a CNN for furniture style classification, testing on their own collected dataset with style labels from industry professionals. A work similar to ours in spirit~\cite{furniture_style}, tries to assess compatibility of furniture, but is based on learning a compatibility function for 3D models and requires computing geometric features for each 3D model. Additionally it doesn't account for material and texture which can carry important information about the style.

\section{Style Compatibility Models}

Our interest in learning about style compatibility of furniture is manifested in two tasks. First and foremost, we would like to be able to assess the style compatibility of two furniture items. We refer to this as the compatibility task. Secondly, given an image of an item of furniture, we would like to be able to retrieve, without knowing its style, other items which are stylistically compatible. Furthermore, we would like to place constraints on the retrieval, \eg~the type of item, or its color, material \etc~ We refer to the second task as the retrieval task.

To tackle these two tasks, we learn Siamese architectures (Sections~\ref{subsec_canonical} to ~\ref{subsec_categorical}) to serve as a stylistic similarity metric between pairs of furniture images. These architectures can directly address the compatibility task. They can also naturally be extended to the retrieval task by making pairwise computations of a query image with respect to a collection of images and returning those which have the highest compatibility score. 
To apply a constrained query, we use using 
joint visual-text embeddings (Section ~\ref{subsec_vse}) for representing both text and images in the same semantic space. 

\begin{figure}[t!]
 \begin{subfigure}[b]{0.4\textwidth}
 \centering
 \includegraphics[width = 0.9\linewidth]{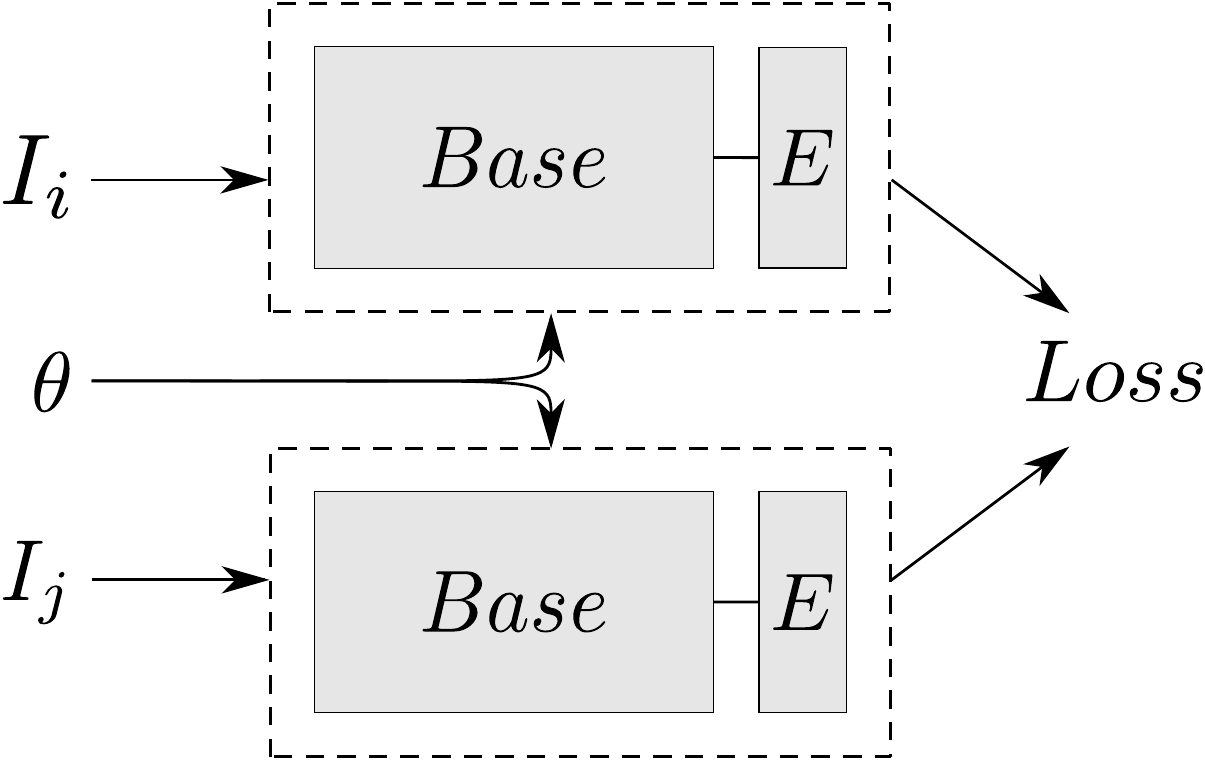}
 \caption{Canonical}
 \label{subfig_canonical}
 \end{subfigure}
 ~
 \begin{subfigure}[b]{0.4\textwidth}
 \centering
 \includegraphics[width = 0.9\linewidth]{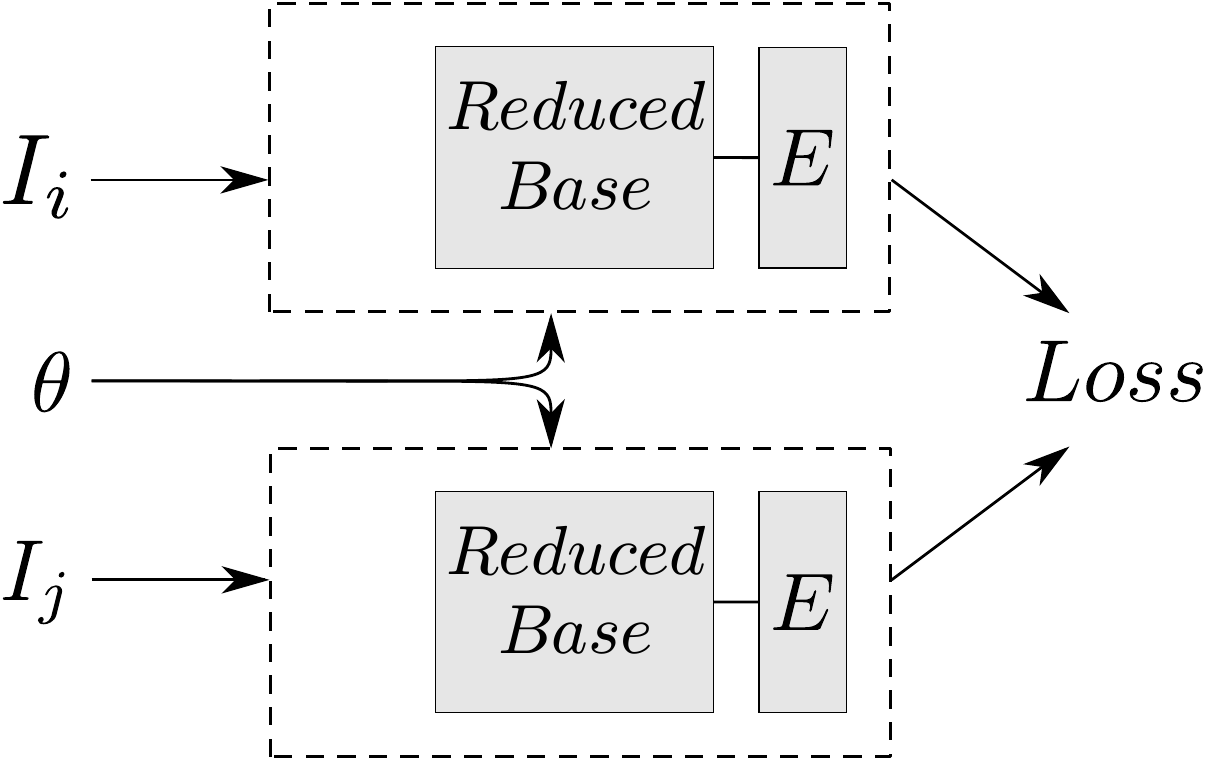}
 \caption{Short}
 \label{subfig_short}
 \end{subfigure}
 ~
 \begin{subfigure}[b]{0.4\textwidth}
 \centering
 \includegraphics[width = 0.9\linewidth]{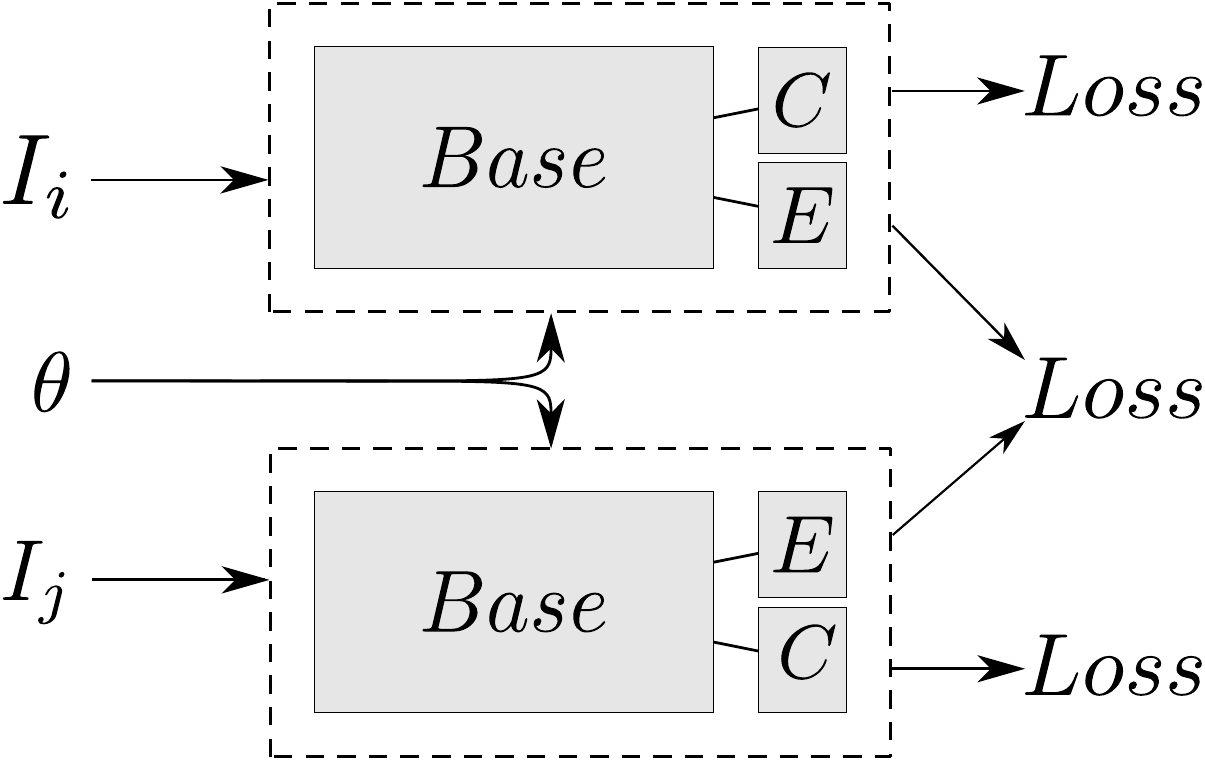}
 \caption{Categorical}
 \label{subfig_categorical}
 \end{subfigure} 
  ~
 \begin{subfigure}[b]{0.4\textwidth}
 \centering
 \includegraphics[width = 0.9\linewidth]{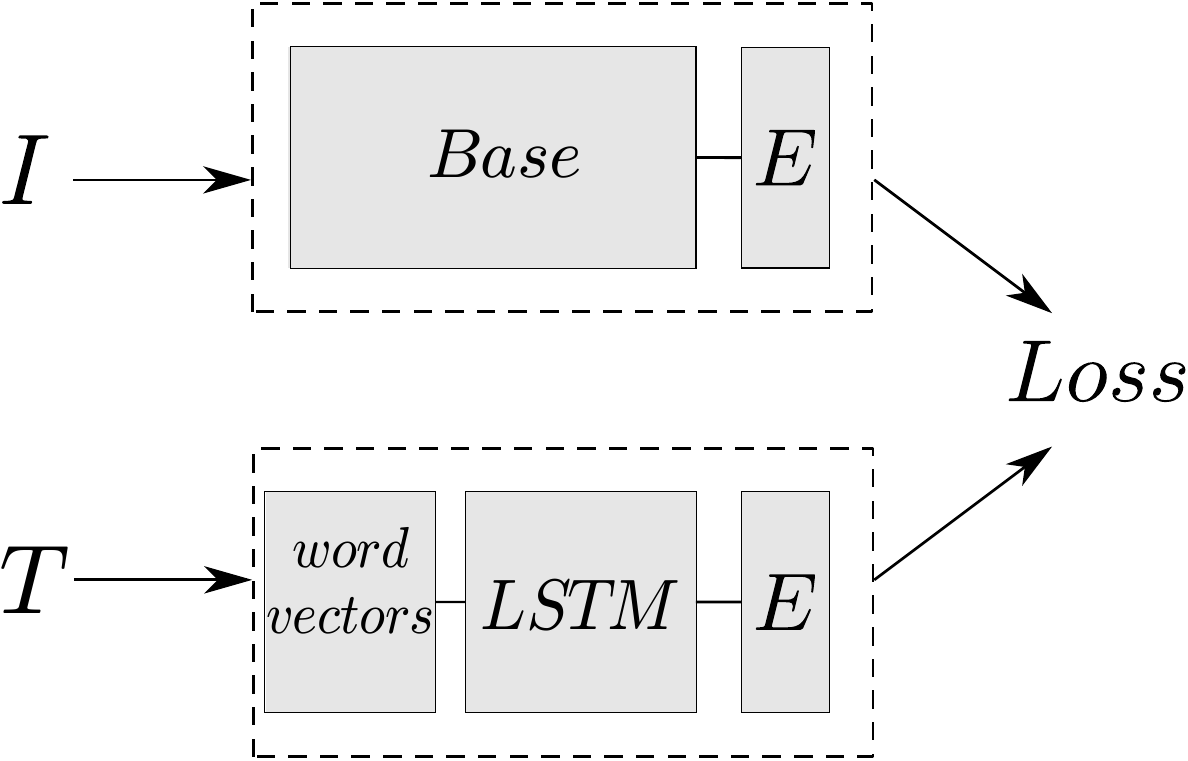}
 \caption{VTE, LSTM.}
 \label{fig_blstm}
 \end{subfigure}
\caption{Structure of our models. $Base$ - base model, $E$ - embedding layer, $C$ - dense layer for classification, $T$ - text as concatenated meta-data, $VTE$ - visual-text embeddings.} 
~\label{fig_siamese_arch}
\end{figure}

\subsection{Canonical Siamese Networks} 
\label{subsec_canonical}
The Siamese architecture joins two sub-networks at their output~\cite{original_siamese} to make some inference about a pair of input data and are commonly used for comparison. Typically, the base of the sub-networks have the same configuration and parameter set. For our \emph{Canonical} Siamese configuration (see Fig.~\ref{subfig_canonical}), we use a pre-trained CNN as the base and add a fully connected layer $E$ with D-dimensional output on top, and apply a contrastive loss to learn an embedding that maps similarly styled furniture items close to each other and different styles further apart:  
\begin{equation}
L_{\text{can}}(I_i,I_j,Y,\theta) = Y\cdot \frac{1}{2} D_\theta(I_i,I_j)^2 + (1-Y) \cdot \frac{1}{2} \max [ 0 , m - D_\theta(I_i,I_j)]^2
\label{eq_con_loss}
\end{equation}
where $(I_i, I_j, Y)$ is a sample pair of images $I_i$ and $I_j$ with compatibility label $Y \in \{0,1\}$ ($Y=1$ for a positive, compatible pair and $Y=0$ for a negative non-compatible pair). $D_\theta(I_i, I_j)$ is a distance measure between $I_i$ and $I_j$. The first term of the loss minimizes the distance between positive image pairs, while the second term of the loss encourages the distance between negative pairs to be as large as possible, up to some margin $m > 0$. Such a margin $m$ is necessary, since it is possible to push negative pairs arbitrarily far from each other and thereby dominate the loss function. Siamese networks maps an image $I_i$ into a D-dimensional embedding location $x_i$. In our experiments, we use the Euclidean distance on these output locations of images $I_i$ and $I_j$ as the distance, \ie~$D_\theta(I_i,I_j) = \|x_i - x_j\|^2$. The margin $m$ is determined via cross-validation. Note that the canonical model does not directly receive any style labels during training; styles are defined implicitly from positive compatible and negative non-compatible pairs.
 
\subsection{Short Siamese Network} 
\label{subsec_short}
Convolutional neural networks learn feature hierarchies, where features increase in semantic complexity as one progresses through the layers~\cite{zeiler2014visualizing}. The mid-network layers have been shown to be important for preserving painting styles when transferring styles for photographs~\cite{gatys2015neural,Johnson2016Perceptual}. Based on this observation, we hypothesize that the later layers of a pretrained CNN are not necessary for differentiating style and experiment with a \emph{Short} Siamese network (see Fig.~\ref{subfig_short}) with only the initial layers of a pre-trained CNN as an alternative base.  
 
\subsection{Categorical Siamese Network}~\label{sec:catsiam}
\label{subsec_categorical}
It has been shown that including a classification loss during the training of Siamese model can improve the performance of image retrieval \cite{similarity,bunny}. To leverage style labels, we combine similarity learning and classification into a multi-task setup and define a \emph{Categorical} Siamese network. This network adds a soft-max layer to the base network so that it produces both an embedding and classification output $C$ (See Fig.~\ref{subfig_categorical}). The loss for the Categorical Siamese is defined as:
\begin{equation}
 L_{\text{cat}}(I_i,C_i,I_j,C_j,Y,\theta) = L_{\text{can}}(I_i,I_j,Y,\theta) + L_{C}(I_i,C_i,\theta) + L_{C}(I_j,C_j,\theta),
\end{equation}
where $L_{\text{can}}$ is the contrastive loss of the canonical network (see Eq.\ref{eq_con_loss}), $C_i$ and $C_j$ are the style annotations, and $L_C$ is a categorical cross-entropy:
\begin{equation}
 L_C = - \frac{1}{N} \sum_{n=1}^N \sum_{i}y_i\log \hat{y}_i,
\label{eq_crossentropy}
\end{equation}
where $y$ is the output of the soft-max layer and $\hat{y}$ is the ground truth.
 
\subsection{Joint Visual-Text Embeddings}~\label{vse_models} 
\label{subsec_vse}
We can further enhance our understanding of style and style compatibility by leveraging the text information from the meta-data of the product images. In particular, we extract meta-data for a product's coloring, material, and furniture type, in addition to its style. It is highly challenging to work with such meta-data because such information is not consistently available for all the images; furthermore it can be of variable length \eg \emph{``cream, wood, dresser''} or \emph{``ruby english tea, hardwood fabric, sofa''}. 

Our objective is to learn a joint embedding space to which both images and text meta-data can be projected (see Fig.~\ref{fig_blstm}). Images and correctly associated meta-data should be kept close together and vice versa for non-matching images and meta-data. We do this in a similar way as DeViSE~\cite{devise}, by projecting both image and text features into a common D-dimensional embedding space, $E$, while minimizing a hinge rank loss: 

\begin{equation}
L_{\text{joint}}(I_i,T_i,\theta) = \sum_{j\neq i} \max[0, m - S_{\theta}(I_i,T_i) + S_{\theta}(I_i,T_j)]
\label{eq_devise_loss}
\end{equation}
where $(I_i, T_i)$ is a sample pair of image $I_i$ and sequence of text $T_i$. Image and text features can be projected into the the embedding positions $x_I$ and $x_T$ in the D-dimensional embedding space, E. Then, $S_\theta(I_i, T_i)$ is the dot product similarity between the vectors $x_I$ and $x_T$ of image $I_i$ and text $T_i$. $T_i$ represents the correct text for image $I_i$, while $T_j$ is the text for images other than $I_i$. This loss encourages correct image and text pairs $(I_i, T_i)$ to have a higher similarity score than the wrong pairs $(I_i, T_j)$ by a margin $m$. 

To represent the meta-data for each product as a fixed-length feature vector (300 in our case), we convert each word in the meta-data to its word2vec~\cite{gensim} representation and apply it as input to a 1-layer Long Short Term Memory network~\cite{hochreiter1997long} (LSTM) with 300 hidden nodes. The corresponding images are passed through a base network and projected into the joint embedding space. 
 
\begin{figure}[b!]
\centering
 \begin{subfigure}[b]{0.19\textwidth}
 \centering
   \includegraphics[width=.4\textwidth]{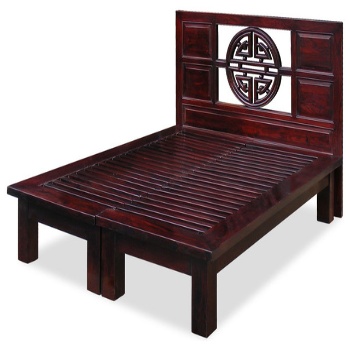}
   \includegraphics[width=.4\textwidth]{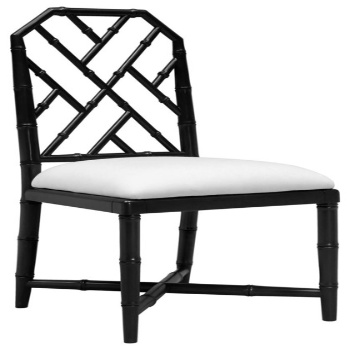}
 \caption {\footnotesize{Asian}}
 \label{subfig_asian}
 \end{subfigure}
 \begin{subfigure}[b]{0.19\textwidth} 
 \centering 
  \includegraphics[width=.4\textwidth]{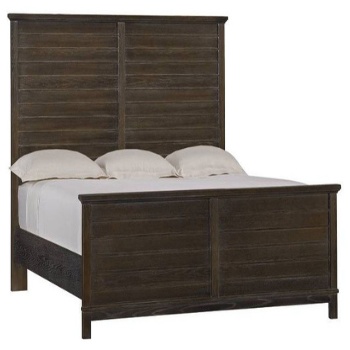}
  \includegraphics[width=.4\textwidth]{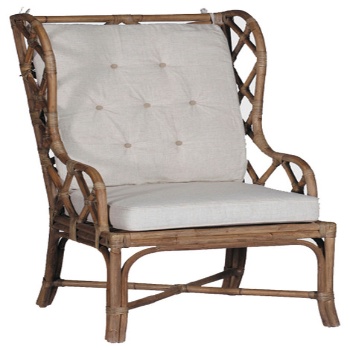}
 \caption{\footnotesize{Beach}}
 \label{subfig_beach}
 \end{subfigure}
 \begin{subfigure}[b]{0.19\textwidth} 
 \centering 
  \includegraphics[width=.4\textwidth]{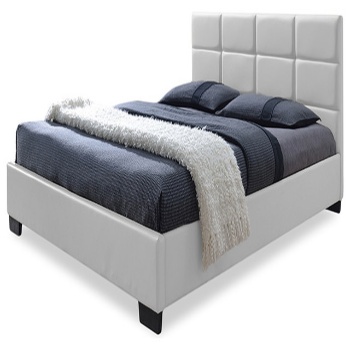}
  \includegraphics[width=.4\textwidth]{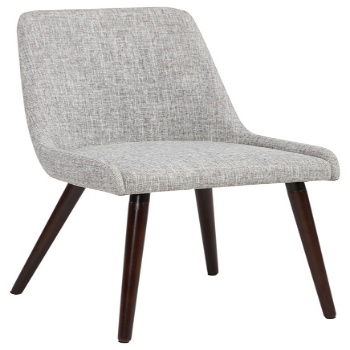}
 \caption{\footnotesize{Contemp.}}
 \label{subfig_contemporary}
 \end{subfigure}
 \begin{subfigure}[b]{0.19\textwidth} 
 \centering 
  \includegraphics[width=.4\textwidth]{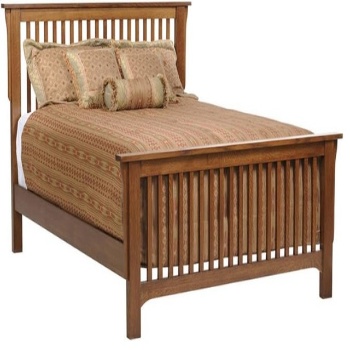}
  \includegraphics[width=.4\textwidth]{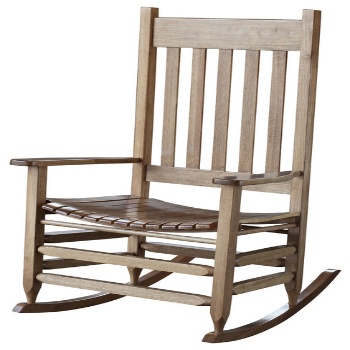}
 \caption{\footnotesize{Craftsman}}
 \label{subfig_craftsman}
 \end{subfigure}
 \begin{subfigure}[b]{0.19\textwidth} 
 \centering 
  \includegraphics[width=.4\textwidth]{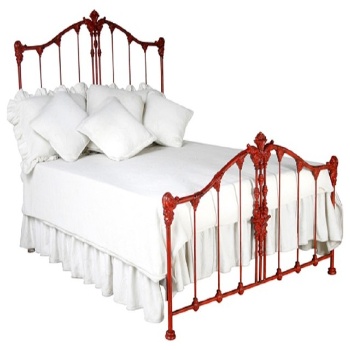}
  \includegraphics[width=.4\textwidth]{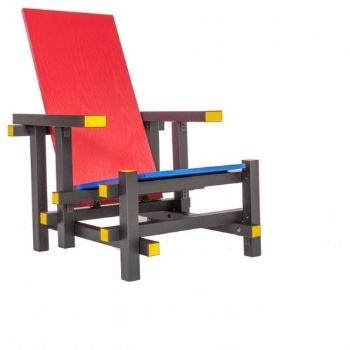}
 \caption{\footnotesize{Eclectic}}
 \label{subfig_eclectic}
 \end{subfigure}
  \begin{subfigure}[b]{0.23\textwidth} 
 \centering 
  \includegraphics[width=.4\textwidth]{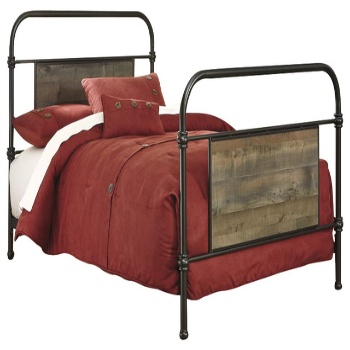}
  \includegraphics[width=.4\textwidth]{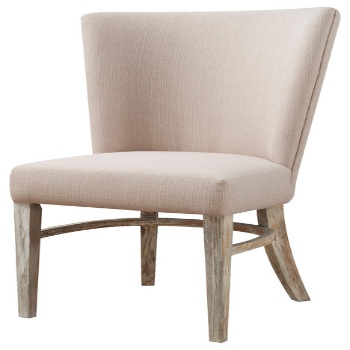}
 \caption{\footnotesize{Farmhouse}}
 \label{subfig_farmhouse}
 \end{subfigure}
  \begin{subfigure}[b]{0.23\textwidth} 
 \centering 
  \includegraphics[width=.4\textwidth]{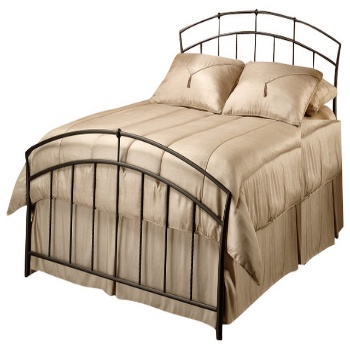}
  \includegraphics[width=.4\textwidth]{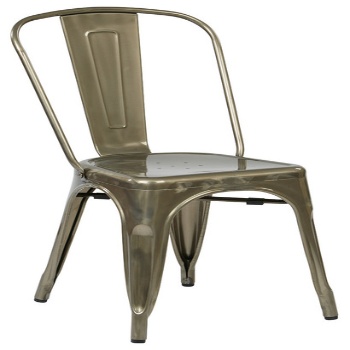}
 \caption{\footnotesize{Industrial}}
 \label{subfig_industrial}
 \end{subfigure}
  \begin{subfigure}[b]{0.23\textwidth} 
 \centering 
  \includegraphics[width=.4\textwidth]{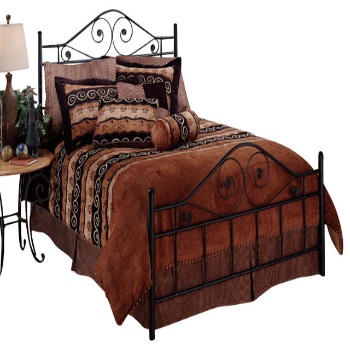}
  \includegraphics[width=.4\textwidth]{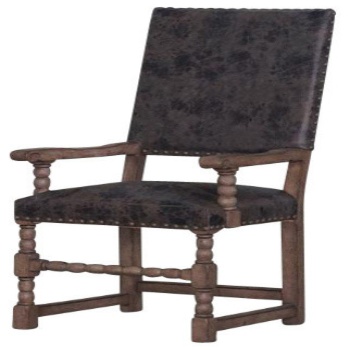}
 \caption{Medit.}
 \label{subfig_mediterranean}
 \end{subfigure}
  \begin{subfigure}[b]{0.23\textwidth} 
 \centering 
  \includegraphics[width=.4\textwidth]{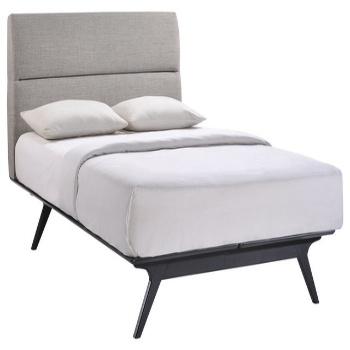}
  \includegraphics[width=.4\textwidth]{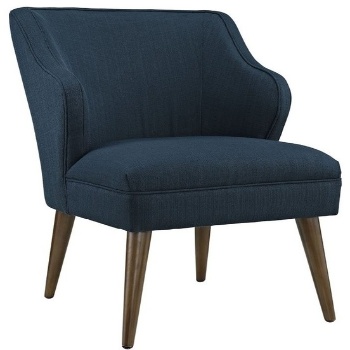}
 \caption{Midcentury}
 \label{subfig_midcentury}
 \end{subfigure}
 \begin{subfigure}[b]{.23\textwidth}
 \centering
  \includegraphics[width=.4\textwidth]{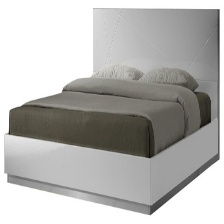}
  \includegraphics[width=.4\textwidth]{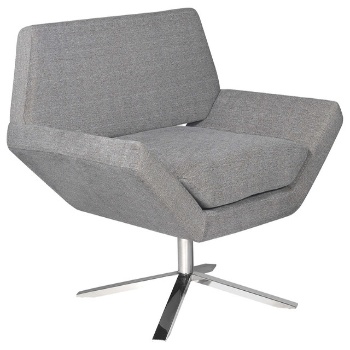}
 \caption{Modern}
 \label{subfig_modern}
 \end{subfigure} 
 \begin{subfigure}[b]{.23\textwidth}
 \centering
  \includegraphics[width=.4\textwidth]{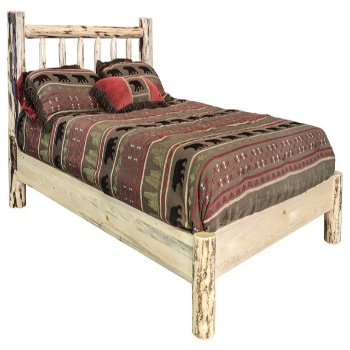}
  \includegraphics[width=.4\textwidth]{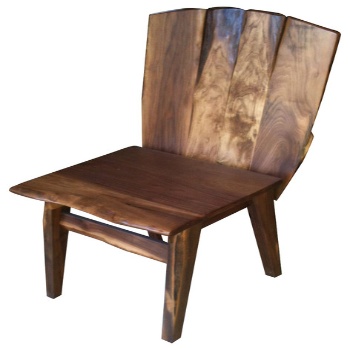}
 \caption{Rustic}
 \label{subfig_rustic}
 \end{subfigure} 
 \begin{subfigure}[b]{.23\textwidth}
 \centering
  \includegraphics[width=.4\textwidth]{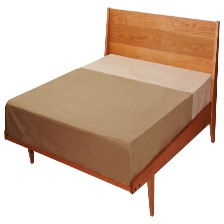}
  \includegraphics[width=.4\textwidth]{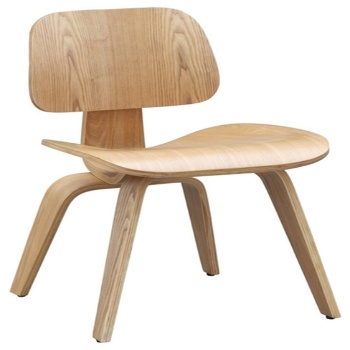}
 \caption{Scandinavian}
 \label{subfig_scandinavian}
 \end{subfigure} 
  \begin{subfigure}[b]{.23\textwidth}
 \centering
  \includegraphics[width=.4\textwidth]{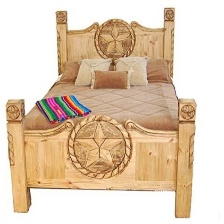}
  \includegraphics[width=.4\textwidth]{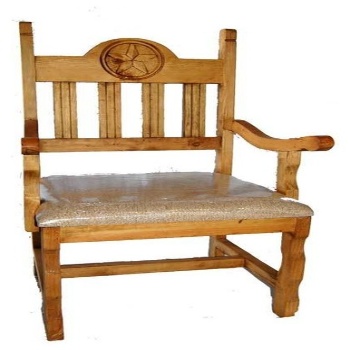}
 \caption{Southwestern}
 \label{subfig_southwestern}
 \end{subfigure}   
 \begin{subfigure}[b]{.23\textwidth}
 \centering
  \includegraphics[width=.4\textwidth]{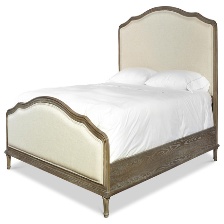}
  \includegraphics[width=.4\textwidth]{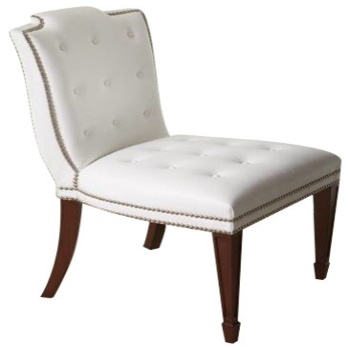}
 \caption{Traditional}
 \label{subfig_traditional}
 \end{subfigure}
 \begin{subfigure}[b]{.23\textwidth}
 \centering
  \includegraphics[width=.4\textwidth]{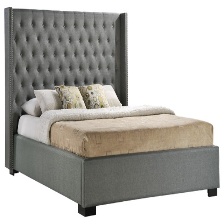}
  \includegraphics[width=.4\textwidth]{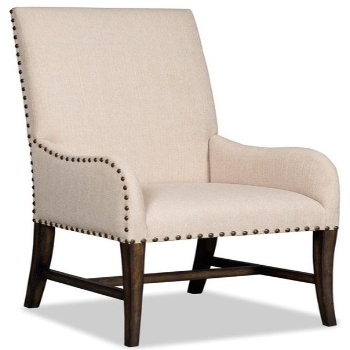}
 \caption{Transitional}
 \label{subfig_transitional}
 \end{subfigure} 
 \begin{subfigure}[b]{.23\textwidth}
 \centering
  \includegraphics[width=.4\textwidth]{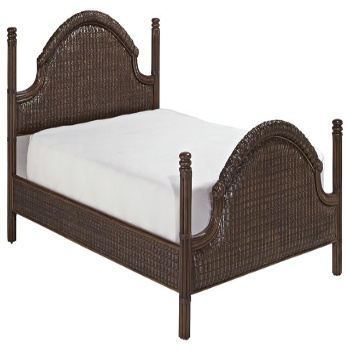}
  \includegraphics[width=.4\textwidth]{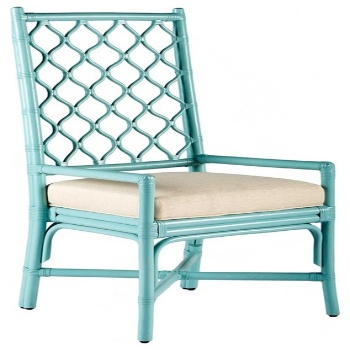}
 \caption{Tropical}
 \label{subfig_tropical}
 \end{subfigure} 
 \begin{subfigure}[b]{.23\textwidth}
 \centering
  \includegraphics[width=.4\textwidth]{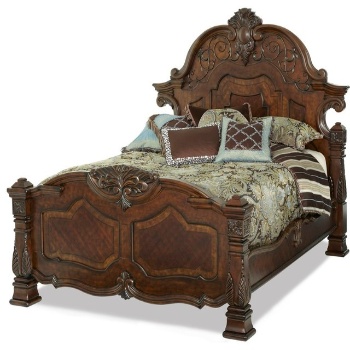}
  \includegraphics[width=.4\textwidth]{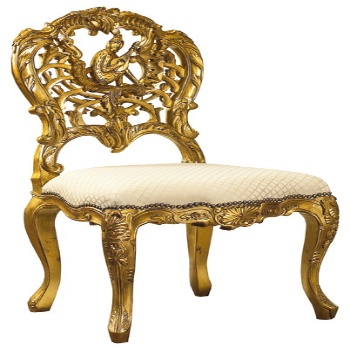}
 \caption{Victorian}
 \label{subfig_victorian}
 \end{subfigure} 
\caption{\emph{``Beds''} and \emph{``chairs''} from the 17 styles. Style similarity may be encoded in material (\ref{subfig_rustic}), geometric curvatures (\ref{subfig_victorian}) or colour (\ref{subfig_eclectic}).}
\label{fig_style_overview}
\end{figure}

\begin{figure}[t!]
\centering
  \includegraphics[width=0.9\linewidth]{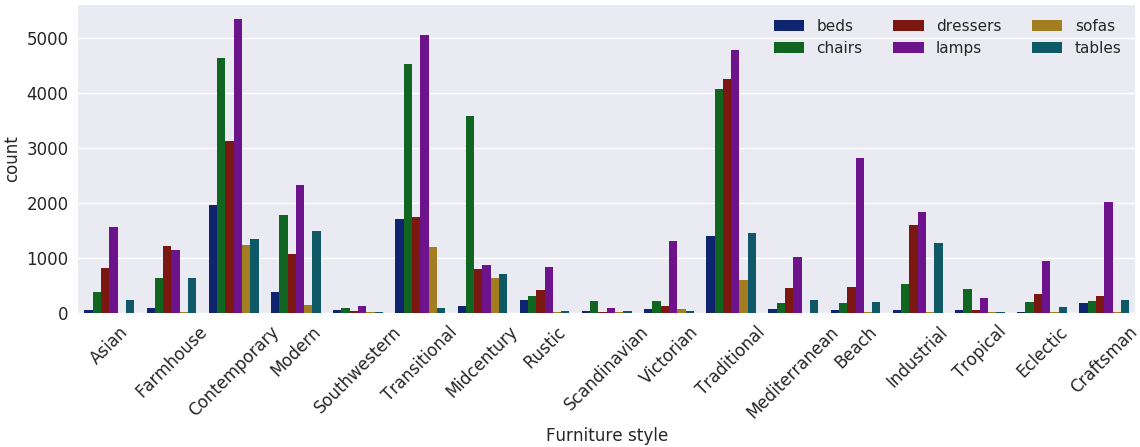}
\caption{Style distribution across categories in the Bonn Furniture Styles Dataset}
\label{fig_style_cat_distribution}
\end{figure}

\section{Bonn Furniture Styles Dataset}~\label{sec_dataset} 
We collect a dataset of 90,298 furniture images with corresponding text meta-data from \texttt{\url{Houzz.com}}, a website specialized in architecture, decoration and interior design. We focus only on iconic images, depicting the furniture on a white background, so as to minimize crosstalk from distractors that may be present in images of the item within a scene. Text meta-data contain information as manufacturer, size, weight, materials, subcategories (e.g folding chair, dining chair, armchair) and, most importantly, the style.
  
Our dataset has the six most popular categories of furniture on \texttt{Houzz.com}: beds (6594), chairs (22247), dressers (16885), lamps (32403), sofas (4080) and tables (8183). The distribution of styles per category are shown in Fig.~\ref{fig_style_cat_distribution}. 
Fig.~\ref{fig_style_overview} provides examples of the styles for the ``bed`` and ``chair'' categories.

Because our dataset and corresponding labels are extracted from the web, it tends to be noisy with mis-labelled data and duplicates. We first remove outliers by fine-tuning GoogLeNet~\cite{googlenet} for the six categories of furniture based on a dataset of 25k images retrieved from the web. Samples with the lowest soft-max scores are considered to be irrelevant and removed. We find duplicates by thresholding on the Hamming distance between computed Perceptual Hash~\cite{phash} values. 
When detecting duplicates, we remove one of the two duplicates if the style annotations are the same; if they are annotated with different styles, we remove both. This step also removes different colored versions of the same product.

\section{Experiments}\label{sec_experiments}

\subsection{Setup, Training Details \& Evaluation}
As a base model, we use a pre-trained GoogLeNet~\cite{googlenet} with the last layer removed. The embedding layer linearly transforms the 1024 dimensional feature vector output from GoogLeNet~\cite{googlenet} into a 256-dimensional embedding feature. For the short Siamese, we first consider the GoogLeNet~\cite{googlenet} layers up to Inception 4c and apply an average pooling that produces a 512-dimensional vector, to which we then append the embedding layer.  For further comparisons, we also experimented with short networks with layers up to Inception 4b, 3b, and 3a.  
 
To generate training pairs, we use a \emph{strategic sampling}~\cite{veit2015learning} that creates positive pairs only from different categories, \ie same style, different types of furniture. This pushes the embedding to focus solely on the style compatibility and not on the similarities between items from the same furniture types. Negative pairs, on the other hand, are pairs of different styles drawn randomly from all types of furniture, both same and different.  

We split the dataset into train, validation and test sets according to a 68:12:20 ratio. Our final training set of pairs has 2.2M pairs, while the validation and test sets have 200k pairs; similar to \cite{veit2015learning}, we keep a 1:16 ratio of positive and negative pairs.  We also test on the Singapore Furniture Dataset~\cite{style_classification}, using provided splits. Due to the smaller size of this dataset, we are able to generate 2.2M pairs for training but only 11K pairs for validation and 65K pairs for testing.
 
All networks are trained in two stages; first we back-propagate only over the embedding / classification layer (with learning rate $\epsilon = 0.01$, weight decay $\rho = 0.9$) for 50 iterations and then do a fine-tuning over the entire network with a lower learning rate ($\epsilon=0.0001$). Overall training takes eight epochs. Embedding / classification soft-max layer weights are randomly initialized. 

For training the visual-text embeddings, we create batches of a reference image and $16$ other randomly sampled images from different styles than the reference. These sets of 17 images, along with their concatenated textual data are then considered as one batch for training the embeddings. The optimizer used for training is RMSProp with a learning rate $\epsilon = 0.001$. The base model for extracting the features from the image is kept fixed, so that only the visual embedding weights is learned. However, for the text part of the model, the entire LSTM as well as the textual embedding weights are learned from scratch.

To determine the margin parameter $m$ in Eq.~\ref{eq_con_loss}, we cross-validate to set a value of $m=50$ for the canonical and short Siamese networks and $m=1000$ for the categorical network.

We evaluate the compatibility task performance with area under the ROC curve (AUC). The Siamese networks, when presented with a pair of images, return a compatibility score between 0 (not compatible at all) and 1 (fully compatible). An AUC value of 0.5 is equal to taking a random guess, while 1.0 indicates perfect discrimination. For the retrieval task, we report recall@K which measures the fraction of times the correct item was found among top K results.  

\begin{figure}
\centering 
 \includegraphics[width = 0.65\linewidth]{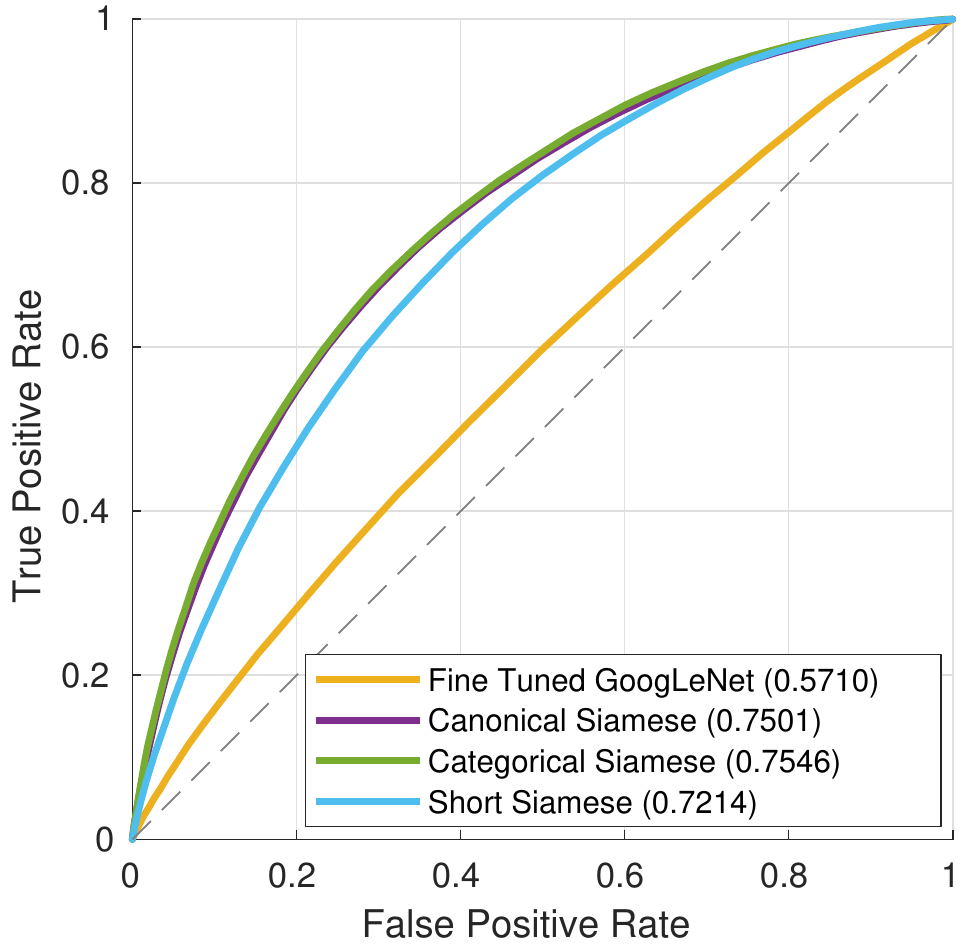}
\caption{Evaluation of models over test pairs generated from our Houzz.com dataset. The plot illustrates ROC curves and corresponding AUC (indicated as ``area''). The dashed line represents random performance (AUC = 0.5).}
\label{roc_dataset}
\end{figure}
 
\subsection{Comparison of Siamese Network Architectures}
\label{comp_base_sia_arch}
First, we compare the performance of three different Siamese network architectures presented in Sections~\ref{subsec_canonical} to~\ref{subsec_categorical}. As a baseline, we fine-tuned the base network GoogLeNet~\cite{googlenet} model to explicitly classify furniture styles, and extracted the CNN features before the classification layer. Later on we compute the Euclidean distance between the CNN features of pairs of images as a measure of similarity. The ROC curves are shown in Figure \ref{roc_dataset}. We find that the canonical and categorical Siamese networks perform very similarly on the compatibility task, with AUCs of 0.7501 and 0.7546 respectively. While the categorical Siamese has only a marginally higher AUC than the canonical, the real impact of the additional categorical cross-entropy loss terms is in the fact that it helps training to converge much faster. After one epoch, the AUC of the categorical Siamese is already 0.732, as opposed to 0.710 for the canonical Siamese.

\begin{table}[]
\begin{tabular}{c|c|c|c|c|c|}
\cline{2-6}
 & Inception 3a & Inception 3b &  Inception 4b & Inception 4c & canonical Siamese \\ \hline
\multicolumn{1}{|c|}{AUC} & 0.6419 & 0.6938 & 0.7201  & 0.7214 & 0.7501 \\ \hline
\multicolumn{1}{|c|}{params} & $\approx0.3M$ &  $\approx0.7M$   & $\approx1.5M$ & $\approx1.9M$ & $\approx6M$ \\ \hline
\end{tabular}
\caption{Evaluation of short Siamese models along with the total number of network parameters. For the short Siamese networks, we consider GoogLeNet~\cite{googlenet} layers up to Inception 3a, 3b, 4b and 4c.}
\label{short_siamese_eval}
\end{table}

More interesting, however, is the performance of the short Siamese network. With an AUC of 0.7214, it performs only a little bit worse than the much deeper canonical version. Such a small gap in performance can be explained by the nature of the stylistic features. Most of the stylistic cues can be visually recognized on a local level, rather than the overall view. For example, the \emph{``Tropic''} style is characterized by woven fabric, whereas \emph{``Victorian''} stands out from specific ornaments and curves. As such, the majority of the relevant features can be captured by the mid-network convolutional layers of a deep network. Given that the short Siamese has much less parameters ($\approx1.9M$ versus $\approx6M$ for the canonical Siamese), it would be more advantageous for resource-limited scenarios such as deployments on a mobile device. In Table~\ref{short_siamese_eval}, we present comparisons for different short Siamese networks along with the number of parameters required.  

\subsubsection{Style-based comparison of all models:} 
\begin{figure}[t!]
\centering
  \includegraphics[width = 0.65\linewidth]{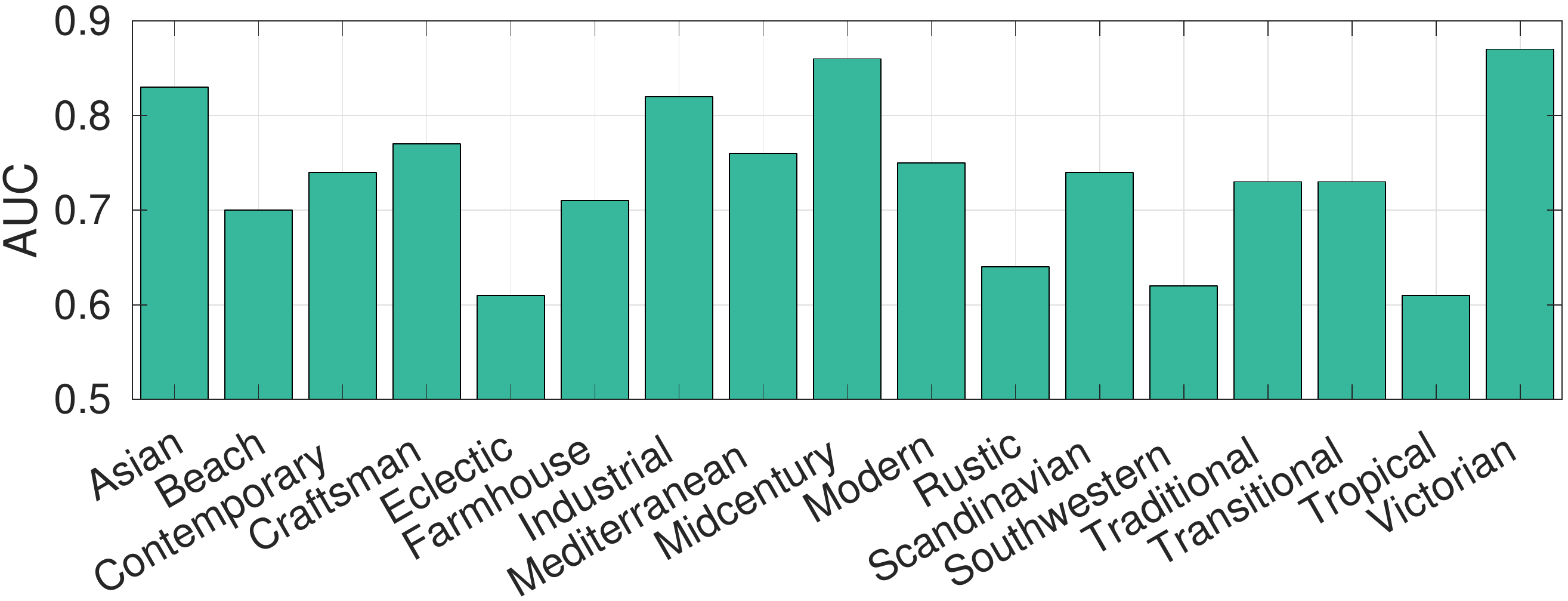}
\caption{AUC scores of the categorical Siamese network on the compatibility task over 17 style classes in the Bonn Furniture Styles Dataset.}
\label{auc_category_based}
\end{figure}

Figure~\ref{auc_category_based} depicts the AUC scores of the categorical Siamese on the compatibility task over the 17 style classes. Among others, \emph{``Victorian''} and \emph{``Midcentury''} styles have the highest AUCs. We attribute this to \emph{``Victorian''} furniture usually having dark wood color with distinct ornaments and sometimes a golden coating. Similarly, \emph{``Midcentury''} furniture can easily be distinguished by inclined legs. In contrast, \emph{``Eclectic''} and \emph{``Southwestern''} styles have the lowest AUCs; the poor performance can be attributed to both a lower number of images in the dataset (see Fig.~\ref{fig_style_cat_distribution}) and large visual variance across furniture types.

\subsection{Style transferability}

\subsubsection{From Compatibility to Classification and Vice Versa:} 
We want to test how well our Siamese networks, trained for style compatibility are able to perform on classifying styles. To do so, we train a linear SVM on the features extracted from the embedding layer of the categorical Siamese model and report our classification results in Table~\ref{tab_acc_compare}. We compare this with a fine-tuned version of the base network, \ie GoogLeNet. We find that our proposed dataset is very difficult; the fine-tuned baseline has a classification accuracy of only 0.41, in comparison to the Singapore dataset~\cite{style_classification}, which has a classification of 0.66. Using the features learned from the embedding of the categorical Siamese, we get better classification performance on our large-scale dataset (0.49), but slightly worse on the smaller Singapore dataset (0.64). In comparison, the best reported classification results on this dataset is is 0.70, achieved through an SVM classification of a combination of fine-tuned CNN features and hand-crafted features~\cite{style_classification}.  Given that the differences to our categorical Siamese is only marginal, we conclude that the features learned by the Siamese network are good representations of the style class.

The reverse case, however, of using features learned for classification for determining style compatibility (see Sec.~\ref{comp_base_sia_arch}), does not hold as well (see Table ~\ref{tab_acc_compare}). As previously reported, the AUC for the categorical Siamese is significantly higher than the fine-tuned GoogLeNet (0.7546 vs 0.5710). We find a similar trend for the Singapore dataset (0.8289 vs 0.6184). Based on these two experiments, we conclude that features learned for style compatibility are well-suited for performing classification but not vice versa.

\subsubsection{Cross-dataset testing:} We investigated the transferability of features learned on one dataset to another by training on our proposed dataset and testing on the Singapore Furniture Dataset~\cite{style_classification}. This dataset has similar types of furniture distributed over 16 styles, 14 of which do not appear in our dataset (only Mediterranean and modern are common to both datasets). We find that the categorical Siamese model trained on our proposed dataset is able to achieve an AUC score of 0.7129 on the Singapore Furniture Dataset~\cite{style_classification}. This suggests that our network is able to generalize to the notion of style similarity, even when encountering never-seen-before styles. When we try the reverse, however, \ie training on the Singapore dataset and testing on ours, the AUC is a much lower 0.5534. We attribute the difference in the performance to the size difference of the two datasets (90K vs. 3K). Although both datasets have almost equal number of style classes, our dataset has more data for each class which allows the deep model to learn finer grained similarities from higher amount of variations within each style across furniture types.
 
\begin{table}[t!]
\centering
\begin{tabular}{|l|c|c|c|c|}
  \hline
  & \multicolumn{2}{c|}{classification}& \multicolumn{2}{c|}{compatibility}\\ \hline
  Model & \multicolumn{1}{l|}{Bonn Styles} & \multicolumn{1}{l|}{Singapore \cite{style_classification}} & \multicolumn{1}{l|}{Bonn Styles} & \multicolumn{1}{l|}{Singapore \cite{style_classification}}\\ \hline
  Fine-tuned GoogLeNet & 0.4141 & \textbf{0.6580} & 0.5710 & 0.6184 \\ \hline
  Categorical Siamese & \textbf{0.4920} & 0.6362 & \textbf{0.7546} & \textbf{0.8289}\\ \hline 
\end{tabular}
\caption{First two columns: comparison of style classification accuracy for a fine-tuned GoogLeNet versus a linear SVM trained on embedding vectors produced by categorical Siamese network. Second two columns: comparison of AUC scores for compatibility assessment between a fine-tuned GoogLeNet versus the categorical Siamese network.}
\label{tab_acc_compare} 
\end{table}

\subsection{Stylistic image retrieval} 
We evaluate the Siamese model's ability to correctly retrieve stylistically similar images in Table~\ref{fig_siamese_recallK}. To evaluate the ranking we report recall@K, $K \in {1,5,10}$. The performance of the different models in the retrieval task mirror that of the compatibility task. The canonical and categorical models perform similarly, with the categorical being slightly better, while the canonical is a little bit lower. In Figure~\ref{siamese_compatibility_search}, we show example retrieval results for the categorical Siamese network. 

\begin{table}[]
\begin{tabular}{c|c|c|c|}
\cline{2-4}
& R@1 & R@5 & R@10 \\ \hline
\multicolumn{1}{|c|}{Canonical}  & 33.7 & 64.1 & 75.6 \\ \hline
\multicolumn{1}{|c|}{Categorical} & 34.5 & 64.8 & 74.9 \\ \hline
\multicolumn{1}{|c|}{Short}    & 29.8 & 63.2 & 76.7 \\ \hline
\end{tabular}
\caption{Stylistic-based retrieval results using Siamese networks. We report recall@K (high is good).}
\label{fig_siamese_recallK}
\end{table}

\subsection{Retrieval with text constraints}
To retrieve stylistically compatible images with additional text constraints, we use the joint visual-text embeddings from Section~\ref{subsec_vse}. We project the query image features extracted from the GoogLeNet to the joint embedding space position $x_I$. Similarly, for the text, which is used as a constraint, we first extract the LSTM~\cite{hochreiter1997long} features and then project them to the joint embedding space position $x_T$. The sum of these two vectors, $x_{sum}=x_I + x_T$, is another position in this space and we return the results based on a nearest-neighbour search in the joint embedding space using $x_{sum}$ as our query vector. 

\begin{figure}[t!]
\centering
 \includegraphics[width = 0.85\linewidth]{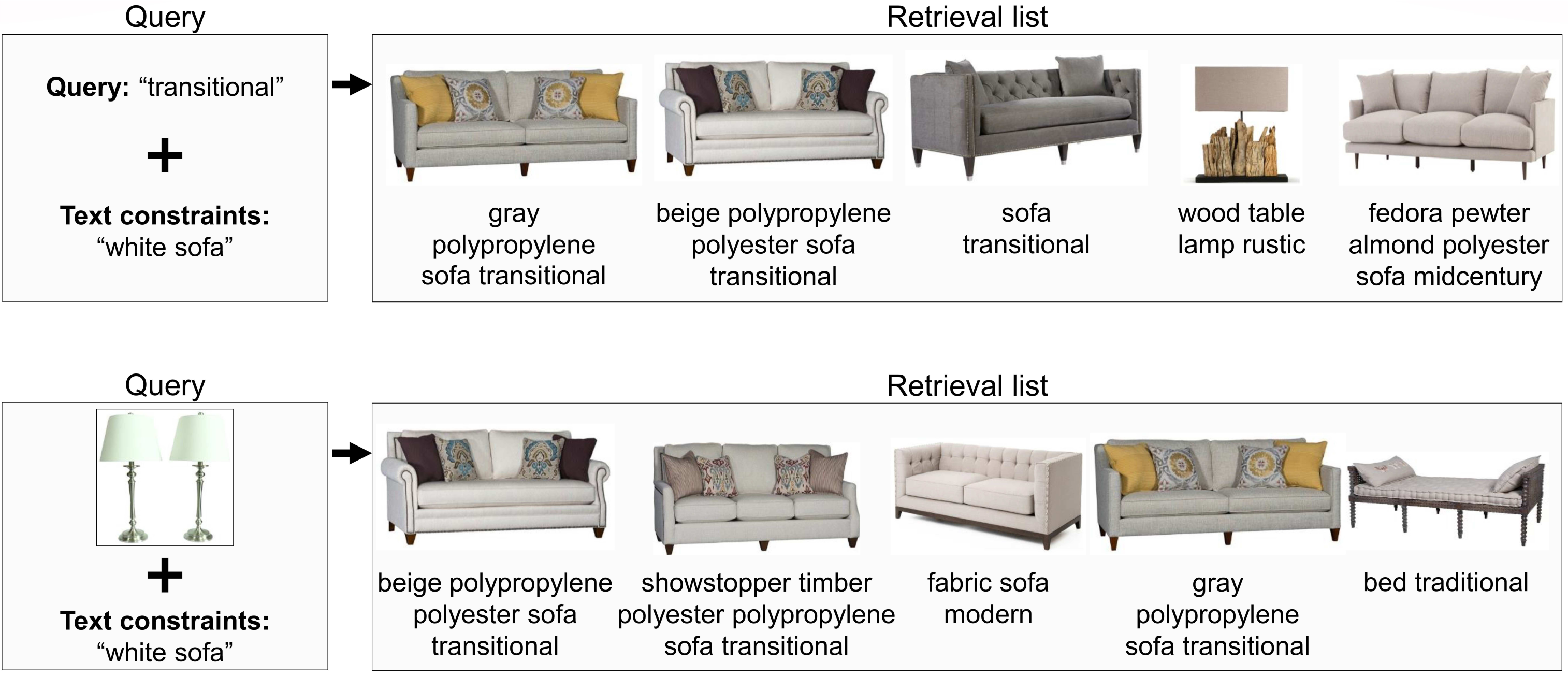}
\caption{Applying our visual text embedding model for queries constrained with text (see Sec.~\ref{subsec_vse}). In the second row, the style of lamps is ``transitional''.}
\label{vse_constraint}
\end{figure}

As a baseline, we use the style text of the query image along with text constraints. We measure the recall of the style and furniture type for such queries and report recall@K, $K \in {1,3,5,10}$ in Table~\ref{fig_devise_recallK} and example retrievals in Figure~\ref{vse_constraint}. Admittedly, our recall performance is not so impressive; however there is only a small drop from querying with images versus text and even the image query is still several times better than chance. One should consider the difficulty of this task, which is expressed qualitatively in Figure~\ref{vse_constraint}: It can be seen that retrieving exact matches is quite difficult, given the available text attributes. Nonetheless, we can still maintain style and category similar to the query image constrained with text.  

\begin{table}[]
\begin{tabular}{c|c|c|c|c|c|c|c|c|}
\cline{2-9}
& \multicolumn{4}{c|}{style+category} & \multicolumn{4}{c|}{exact match} \\ \cline{2-9} 
& R@1   & R@3  & R@5  & R@10  & R@1  & R@3  & R@5  & R@10  \\ \hline
\multicolumn{1}{|c|}{text query}                        
& 0.8   & 1.9  & 4.9  & 6.9   & 0.3  & 0.3  & 0.6  & 0.8  \\ \hline
\multicolumn{1}{|c|}{\begin{tabular}[c]{@{}c@{}}image query\end{tabular}} 
& 0.1   & 1.2  & 3.0  & 5.9   & 0   & 0   & 0.2  & 0.7  \\ \hline
\end{tabular}
\caption{Retrieval with text constraints. We report recall@K (high is good)}
~\label{fig_devise_recallK}
\end{table}

\section{Conclusion}
\label{chapter_conclusion}

In this work we tackle the problem of learning style compatibility for furniture and addressed two different style tasks -- one assessing compatibility between pairs of images, and one querying for stylistically compatible images. To accomplish this, we propose and test several Siamese architectures.
Our evaluations show that Siamese networks excel at capturing the stylistic compatibility; in addition, these networks can be reduced to only low- and mid-level layers of pretrained CNNs and still perform comparably with full CNNs, thereby enabling efficient deployments. In addition, we use a joint visual-text embedding model for making stylistically compatible recommendations, while constraining the queries by attributes like furniture type, color and material. For evaluation, we collected an extensive dataset of furniture images of different style categories, along with textual annotations which we will release publicly. We report comprehensive evaluations on our dataset and show that it is very beneficial for style understanding with and without text.
 
\bibliographystyle{splncs04}
\bibliography{egbib}

\begin{thebibliography}{10}
\providecommand{\url}[1]{\texttt{#1}}
\providecommand{\urlprefix}{URL }
\providecommand{\doi}[1]{https://doi.org/#1}

\bibitem{similarity}
Bell, S., Bala, K.: Learning visual similarity for product design with
  convolutional neural networks. ACM Trans. Graph.  (2015)

\bibitem{original_siamese}
Bromley, J., Guyon, I., LeCun, Y., S\"{a}ckinger, E., Shah, R.: Signature
  verification using a "siamese" time delay neural network. In: Cowan, J.D.,
  Tesauro, G., Alspector, J. (eds.) Advances in Neural Information Processing
  Systems 6, pp. 737--744. Morgan-Kaufmann (1994),
  \url{http://papers.nips.cc/paper/769-signature-verification-using-a-siamese-time-delay-neural-network.pdf}

\bibitem{datta}
Datta, R., Joshi, D., Li, J., Wang, J.Z.: Studying aesthetics in photographic
  images using a computational approach. In: Proceedings of the 9th European
  Conference on Computer Vision - Volume Part III. ECCV'06 (2006)

\bibitem{interestingness}
Dhar, S., Ordonez, V., Berg, T.L.: High level describable attributes for
  predicting aesthetics and interestingness. In: Proceedings of the 2011 IEEE
  Conference on Computer Vision and Pattern Recognition. CVPR '11 (2011)

\bibitem{doersch2012makes}
Doersch, C., Singh, S., Gupta, A., Sivic, J., Efros, A.: {What makes Paris look
  like Paris?} ACM Transactions on Graphics  \textbf{31}(4) (2012)

\bibitem{devise}
Frome, A., Corrado, G.S., Shlens, J., Bengio, S., Dean, J., Ranzato, M.A.,
  Mikolov, T.: Devise: A deep visual-semantic embedding model. In: Advances in
  Neural Information Processing Systems 26 (2013)

\bibitem{illustration_style}
Garces, E., Agarwala, A., Gutierrez, D., Hertzmann, A.: A similarity measure
  for illustration style. ACM Trans. Graph.  (2014)

\bibitem{gatys2015neural}
Gatys, L.A., Ecker, A.S., Bethge, M.: A neural algorithm of artistic style.
  arXiv preprint arXiv:1508.06576  (2015)

\bibitem{architecture}
Goel, A., Juneja, M., Jawahar, C.V.: Are buildings only instances?: Exploration
  in architectural style categories. In: Proceedings of the Eighth Indian
  Conference on Computer Vision, Graphics and Image Processing. ICVGIP '12
  (2012)

\bibitem{bilstm}
Han, X., Wu, Z., Jiang, Y., Davis, L.S.: Learning fashion compatibility with
  bidirectional lstms. CoRR  (2017)

\bibitem{hochreiter1997long}
Hochreiter, S., Schmidhuber, J.: Long short-term memory. Neural computation
  \textbf{9}(8),  1735--1780 (1997)

\bibitem{style_classification}
Hu, Z., Wen, Y., Liu, L., Jiang, J., Hong, R., Wang, M., Yan, S.: Visual
  classification of furniture styles. ACM Trans. Intell. Syst. Technol.  (2017)

\bibitem{memorable}
Isola, P., Xiao, J., Parikh, D., Torralba, A., Oliva, A.: What makes a
  photograph memorable? IEEE Trans. Pattern Anal. Mach. Intell.  (2014)

\bibitem{Johnson2016Perceptual}
Johnson, J., Alahi, A., Fei-Fei, L.: Perceptual losses for real-time style
  transfer and super-resolution. In: European Conference on Computer Vision
  (2016)

\bibitem{karayev2013recognizing}
Karayev, S., Trentacoste, M., Han, H., Agarwala, A., Darrell, T., Hertzmann,
  A., Winnemoeller, H.: Recognizing image style. arXiv preprint arXiv:1311.3715
   (2013)

\bibitem{painter}
Keren, D.: Painter identification using local features and naive bayes. In:
  Proc. of Pattern Recognition Conference. IEEE (2002)

\bibitem{wherebuy}
Kiapour, M.H., Han, X., Lazebnik, S., Berg, A.C., Berg, T.L.: Where to buy it:
  Matching street clothing photos in online shops. In: Proceedings of the 2015
  International Conference on Computer Vision. IEEE Press (2015)

\bibitem{kiapour2014hipster}
Kiapour, M.H., Yamaguchi, K., Berg, A.C., Berg, T.L.: Hipster wars: Discovering
  elements of fashion styles. In: European conference on computer vision (2014)

\bibitem{phash}
Krawetz, N.: Looks like it (2011),
  \url{http://www.hackerfactor.com/blog/index.php?/
  archives/432-Looks-Like-It.html}

\bibitem{aesthetic}
Li, C., Chen, T.: Aesthetic visual quality assessment of paintings. IEEE
  Journal of Selected Topics in Signal Processing  (2009)

\bibitem{lin2015learning}
Lin, T.Y., Cui, Y., Belongie, S., Hays, J.: Learning deep representations for
  ground-to-aerial geolocalization. In: Proceedings of the IEEE conference on
  computer vision and pattern recognition (2015)

\bibitem{furniture_style}
Liu, T., Hertzmann, A., Li, W., Funkhouser, T.: Style compatibility for 3d
  furniture models. ACM Trans. Graph.  (2015)

\bibitem{mcauley2015image}
McAuley, J., Targett, C., Shi, Q., Van Den~Hengel, A.: Image-based
  recommendations on styles and substitutes. In: Proceedings of the 38th
  International ACM SIGIR Conference on Research and Development in Information
  Retrieval. ACM (2015)

\bibitem{gensim}
{\v R}eh{\r u}{\v r}ek, R., Sojka, P.: Software framework for topic modelling
  with large corpora. In: {Proceedings of the LREC 2010 Workshop on New
  Challenges for NLP Frameworks}. ELRA (2010)

\bibitem{bunny}
Sangkloy, P., Burnell, N., Ham, C., Hays, J.: The sketchy database: Learning to
  retrieve badly drawn bunnies. ACM Transactions on Graphics (proceedings of
  SIGGRAPH)  (2016)

\bibitem{art}
Shamir, L., Macura, T., Orlov, N., Eckley, D.M., Goldberg, I.G.: Impressionism,
  expressionism, surrealism: Automated recognition of painters and schools of
  art. ACM Trans. Appl. Percept.  (2010)

\bibitem{neuroaesthetics}
Simo-Serra, E., Fidler, S., Moreno-Noguer, F., Urtasun, R.: Neuroaesthetics in
  fashion: Modeling the perception of fashionability. In: Proc. Computer Vision
  and Pattern Recognition Conference (CVPR'15). IEEE Press (2015)

\bibitem{fashion_128}
Simo-Serra, E., Ishikawa, H.: Fashion style in 128 floats: Joint ranking and
  classification using weak data for feature extraction. In: Proceedings of the
  Conference on Computer Vision and Pattern Recognition (CVPR) (2016)

\bibitem{googlenet}
Szegedy, C., Liu, W., Jia, Y., Sermanet, P., Reed, S., Anguelov, D., Erhan, D.,
  Vanhoucke, V., Rabinovich, A.: Going deeper with convolutions. In: Computer
  Vision and Pattern Recognition (CVPR) (2015)

\bibitem{szegedy2016rethinking}
Szegedy, C., Vanhoucke, V., Ioffe, S., Shlens, J., Wojna, Z.: Rethinking the
  inception architecture for computer vision. In: Proceedings of the IEEE
  conference on computer vision and pattern recognition (2016)

\bibitem{veit2015learning}
Veit, A., Kovacs, B., Bell, S., McAuley, J., Bala, K., Belongie, S.: Learning
  visual clothing style with heterogeneous dyadic co-occurrences. In:
  Proceedings of the IEEE International Conference on Computer Vision (2015)

\bibitem{wang2014learning}
Wang, J., Song, Y., Leung, T., Rosenberg, C., Wang, J., Philbin, J., Chen, B.,
  Wu, Y.: Learning fine-grained image similarity with deep ranking. In:
  Proceedings of the IEEE Conference on Computer Vision and Pattern Recognition
  (2014)

\bibitem{zeiler2014visualizing}
Zeiler, M.D., Fergus, R.: Visualizing and understanding convolutional networks.
  In: European conference on computer vision (2014)

\end{thebibliography}


\begin{thebibliography}{1}
\providecommand{\url}[1]{\texttt{#1}}
\providecommand{\urlprefix}{URL }
\providecommand{\doi}[1]{https://doi.org/#1}

\bibitem{style_classification}
Hu, Z., Wen, Y., Liu, L., Jiang, J., Hong, R., Wang, M., Yan, S.: Visual
  classification of furniture styles. ACM Trans. Intell. Syst. Technol.  (2017)

\bibitem{tsne}
van~der Maaten, L., Hinton, G.: Visualizing data using t-sne. Journal of
  Machine Learning Research  \textbf{9},  2579--2605 (2012)

\bibitem{googlenet}
Szegedy, C., Liu, W., Jia, Y., Sermanet, P., Reed, S., Anguelov, D., Erhan, D.,
  Vanhoucke, V., Rabinovich, A.: Going deeper with convolutions. In: Computer
  Vision and Pattern Recognition (CVPR) (2015)

\end{thebibliography}

\end{document}


\title{Learning Style Compatibility for Furniture Supplementary} 

\titlerunning{Learning Style Compatibility for Furniture Supplementary}

\author{Divyansh Aggarwal\inst{1} \and
Elchin Valiyev\inst{2} \and
Fadime Sener\inst{2} \and
Angela Yao\inst{2} 
}
%
\authorrunning{D. Aggarwal, E. Valiyev, F. Sener, A. Yao}

\institute{IIT Jodhpur, India \and
University of Bonn, Germany\\ 
\email{aggarwal.1@iitj.ac.in}\\ 
\email{s6elvali@uni-bonn.de, \{sener,yao\}@cs.uni-bonn.de}}
\maketitle              
\section{Bonn Furniture Styles Dataset}
\begin{figure} 
\centering
\includegraphics[width=\linewidth]{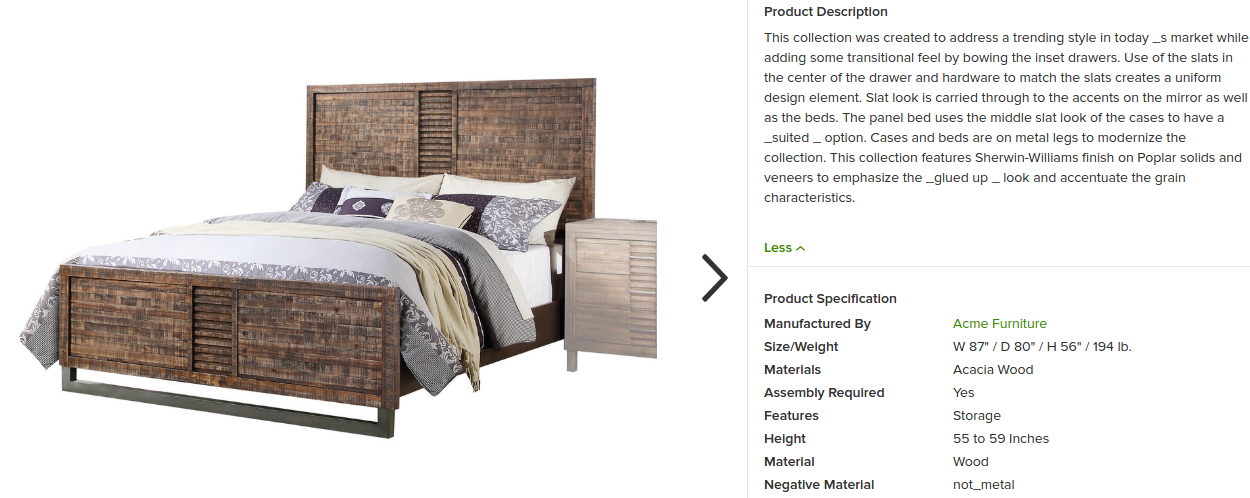}
\caption{A \emph{``rustic''} bed with product description and specifications from \texttt{Houzz.com}.}
\label{fig_dataset_example}
\end{figure}

\begin{figure}[t!]
 \begin{subfigure}[b]{\textwidth}
    \centering
    \includegraphics[width=\linewidth]{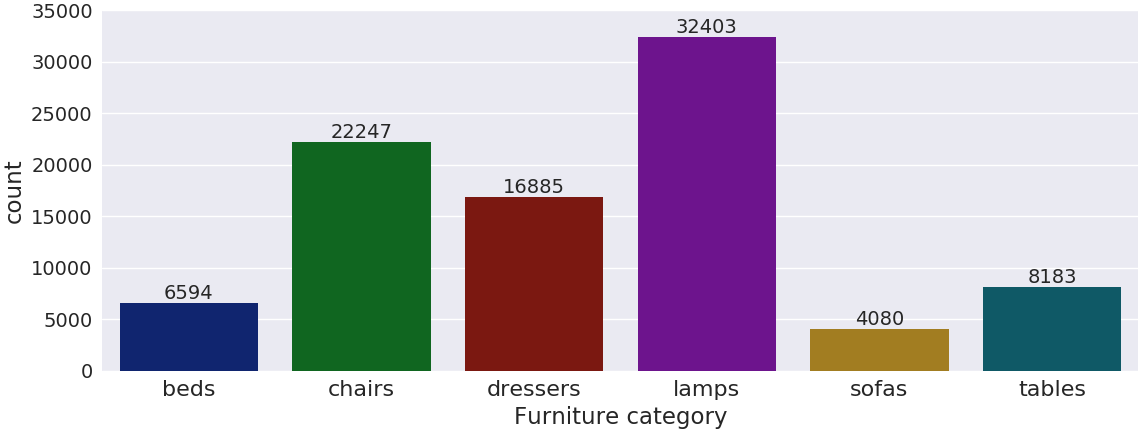}
    \caption{Distribution of furniture types in the dataset.}
    \label{fig_style_cat_sizes}
 \end{subfigure}
 ~
   \begin{subfigure}[b]{\textwidth}
    \centering
    \includegraphics[width=\linewidth]{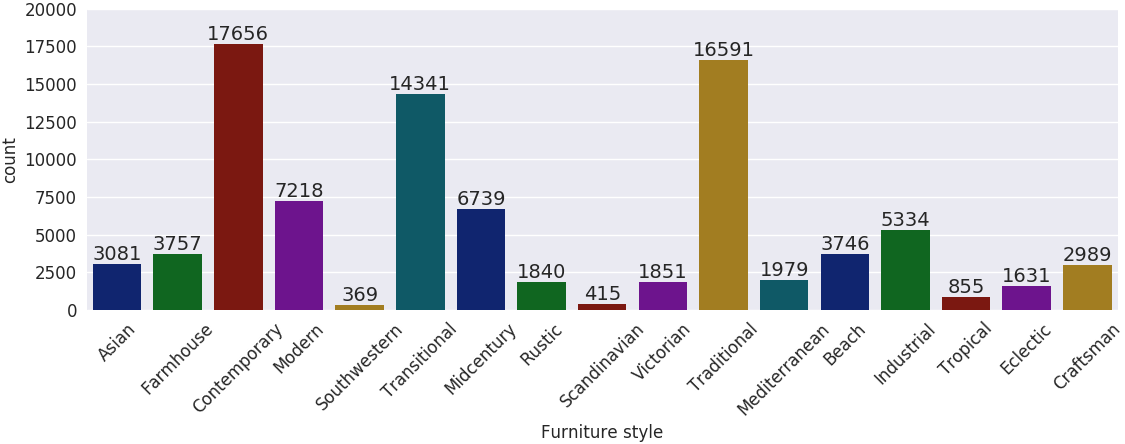}
    \caption{Style distribution for the entire dataset.}
    \label{fig_style_distribution}
 \end{subfigure}  
 ~
  \begin{subfigure}[b]{\textwidth}
    \centering
    \includegraphics[width=\linewidth]{img/dataset_info/cat_style_stat.png}
    \caption{Style distribution across different furniture types in the dataset.}
    \label{fig_style_cat_distribution}
 \end{subfigure}
\caption{Bonn Furniture Styles Dataset statistics.} 
~\label{fig_style_all}
\end{figure}

We collected our dataset from \texttt{Houzz.com}, a website specialized in architecture, decoration and interior design. Our dataset contains images and text descriptions of furniture (See Fig.~\ref{fig_dataset_example}).  We focus only on iconic images of products that depict the items on a white background.  We collected images for the six most popular types of furniture on \texttt{Houzz.com}: beds (6594), chairs (22247), dressers (16885), lamps (32403), sofas (4080) and tables (8183) (See Fig.\ref{fig_style_cat_sizes}).  The distribution of styles over the dataset and across different furniture types are shown in  Fig.~\ref{fig_style_distribution} and Fig.~\ref{fig_style_cat_distribution} respectively.

\section{Experiments}

\subsection{Margin choice} 
Table~\ref{tab_margins} shows the AUC scores of the categorical Siamese networks for varying margin values, $m$, for the Bonn Furniture Styles Dataset and Singapore Furniture Dataset~\cite{style_classification}. The scores are computed on the validation sets. Although the trained model is still able to separate positive and negative pairs, margin $m=1$ produces low results on both datasets. Additionally, using a very large margin, $m=10,000,000$, leads to divergence as negative pairs always contribute to the loss and disrupt the model's convergence to a correct mapping.  

 \begin{table}
 \centering
 \begin{tabular}{|c|c|c|c|c|c|}
   \hline
   Dataset & m=1 & m=10 & m=1,000 & m=3000 & m=10,000,000 \\ \hline
   Singapore \cite{style_classification} & 0.7433 & 0.8738 & 0.8570 & - & 0.5783 \\ \hline
   Bonn Styles  & 0.7221 & 0.7550 & 0.7625 & 0.7103 & 0.5654 \\ \hline
 \end{tabular}
 \caption{AUC scores for different margin values, $m$.  }
 \label{tab_margins}
 \end{table}
 
\begin{figure}
 \centering
   \begin{subfigure}[t]{0.3\textwidth}
   \centering
     \includegraphics[width=\textwidth]{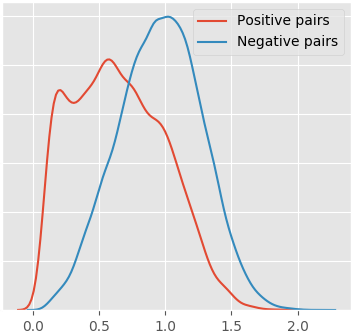}
   \caption{Margin = 1, AUC = 0.74}
   \label{subfig_low_margin}
   \end{subfigure}
   ~ 
   \begin{subfigure}[t]{0.3\textwidth}
     \includegraphics[width=\textwidth]{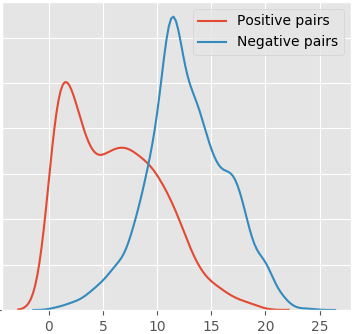} 
   \caption{Margin = 10, AUC = 0.8736}
   \label{subfig_good_margin}
   \end{subfigure}
   ~ 
   \begin{subfigure}[t]{0.3\textwidth}
     \includegraphics[width=\textwidth]{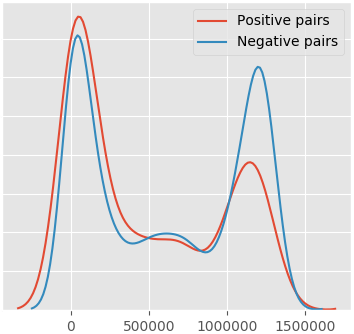}
   \caption{Margin = 10,000,000, AUC = 0.5783}
   \label{subfig_high_margin}
   \end{subfigure}
 \caption{Influence of margin on the distance distributions of the positive and negative pairs. Results are reported on the validation set of the Singapore Furniture Dataset~\cite{style_classification}. Plots are the distance distributions for the positive (red) and negative (blue) pairs.}
 \label{fig_margins}
 \end{figure}

Fig.~\ref{fig_margins} illustrates the influence of margin on the distance distributions of the positive and negative pairs on the Singapore Furniture Dataset~\cite{style_classification}. We report our results on the validation set using a categorical Siamese network. We first compute the distances between the embedding vectors of each negative and positive pair, and then plot the univariate kernel density estimates. Plots are the distance distributions for the positive (red) and negative (blue) pairs. Our models are learning to minimize the overlap between the two distributions. Fig.~\ref{subfig_good_margin} shows that margin $m=10$ produces better separation (AUC=0.8736) compared to margins $m=1$ (AUC=0.74) and $m=10,000,000$ (AUC=0.5783). Using a very large margin leads to almost complete overlap of the distributions and therefore makes the model unable to differentiate the positive and negative pairs (See Fig.~\ref{subfig_high_margin}). In our experiments, we determine the margin $m$ via cross-validation.

\subsection{T-SNE visualizations: }
\begin{figure}
  \begin{subfigure}[t]{0.47\textwidth}
  \centering
    \includegraphics[height=0.8\linewidth] {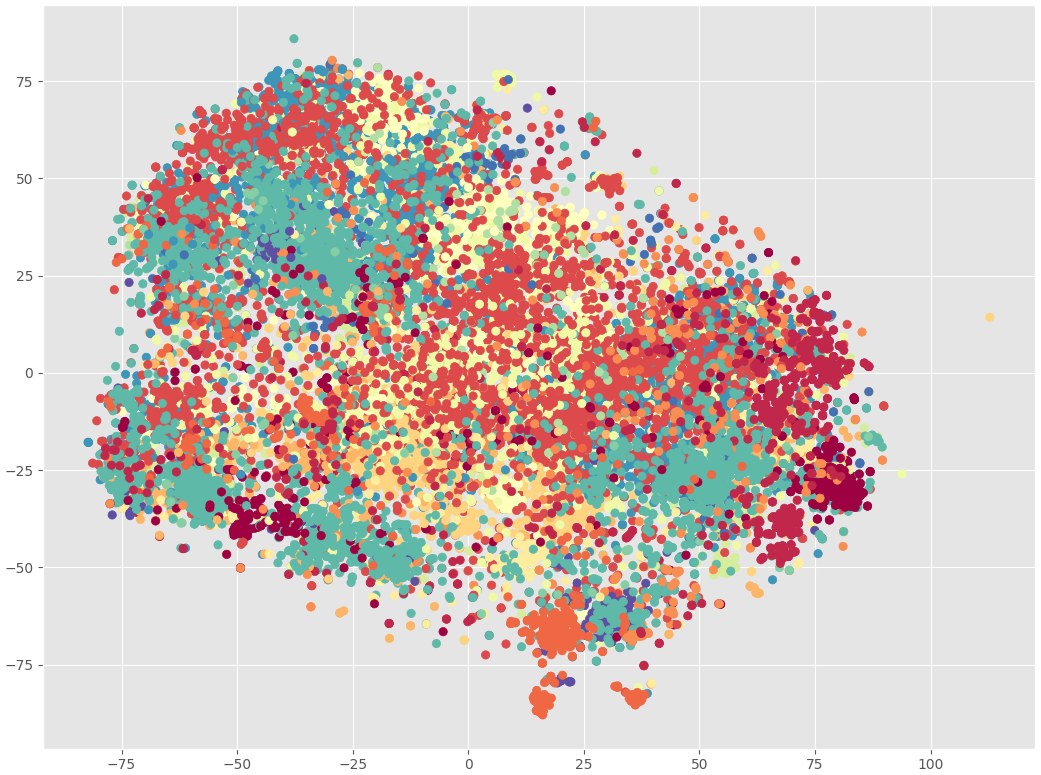}
  \caption{Bonn Styles, GoogLeNet.}
  \label{subfig_tsne}
  \end{subfigure}
  ~ 
  \begin{subfigure}[t]{0.47\textwidth}
  \centering
    \includegraphics[height = 0.8\linewidth] {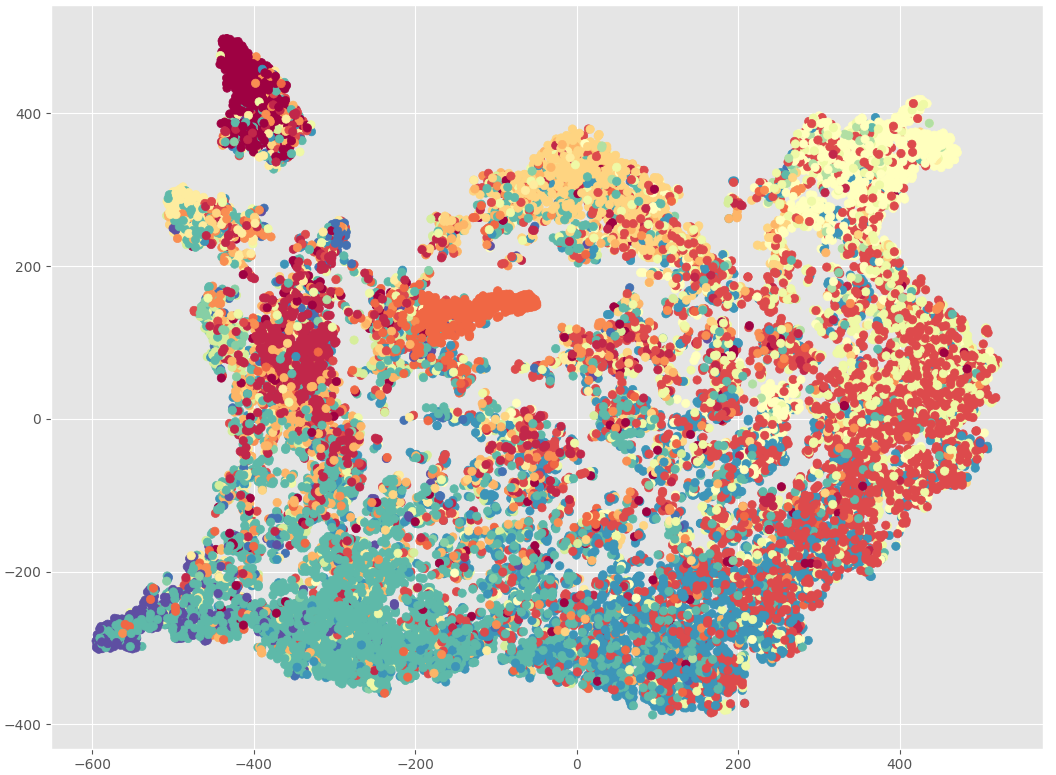}
  \caption{Bonn Styles, Categorical.}
  \label{subfig_tsne_siam}
  \end{subfigure}
  ~ 
   \begin{subfigure}[b]{0.47\textwidth}
   \centering
     \includegraphics[height=0.8\linewidth]{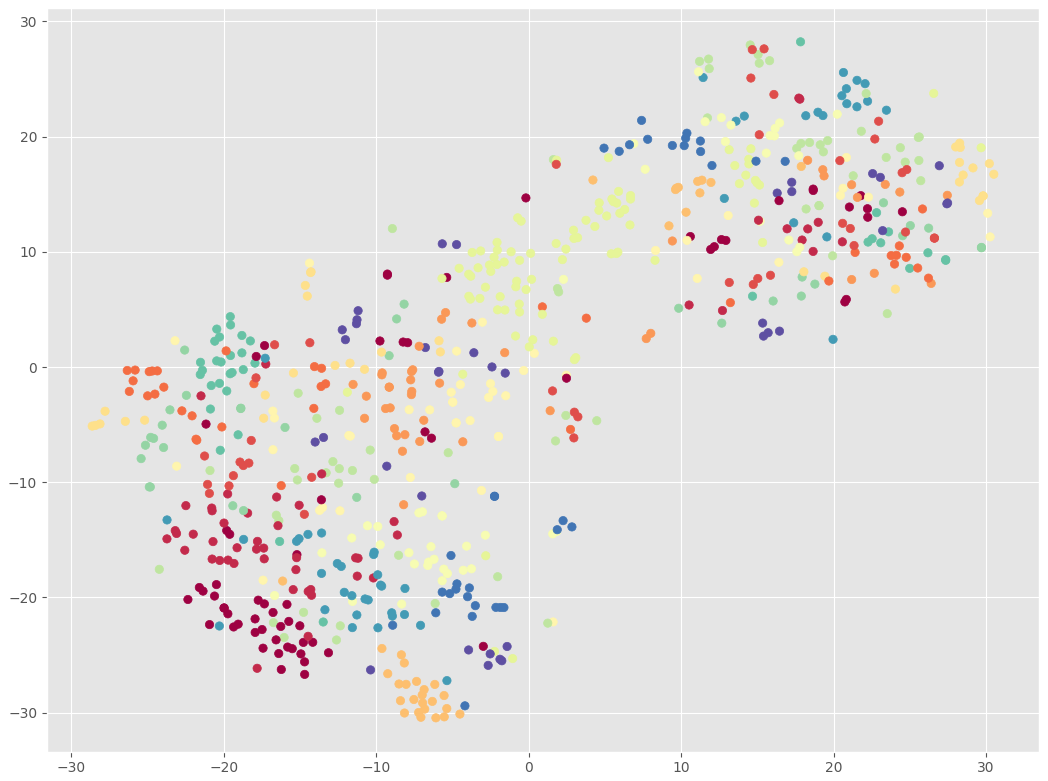}
   \caption{Singapore~\cite{style_classification}, GoogLeNet.}
   \label{subfig_tsne_hu}
   \end{subfigure}
   ~ 
   \begin{subfigure}[b]{0.47\textwidth}
   \centering
     \includegraphics[height = 0.8\linewidth] {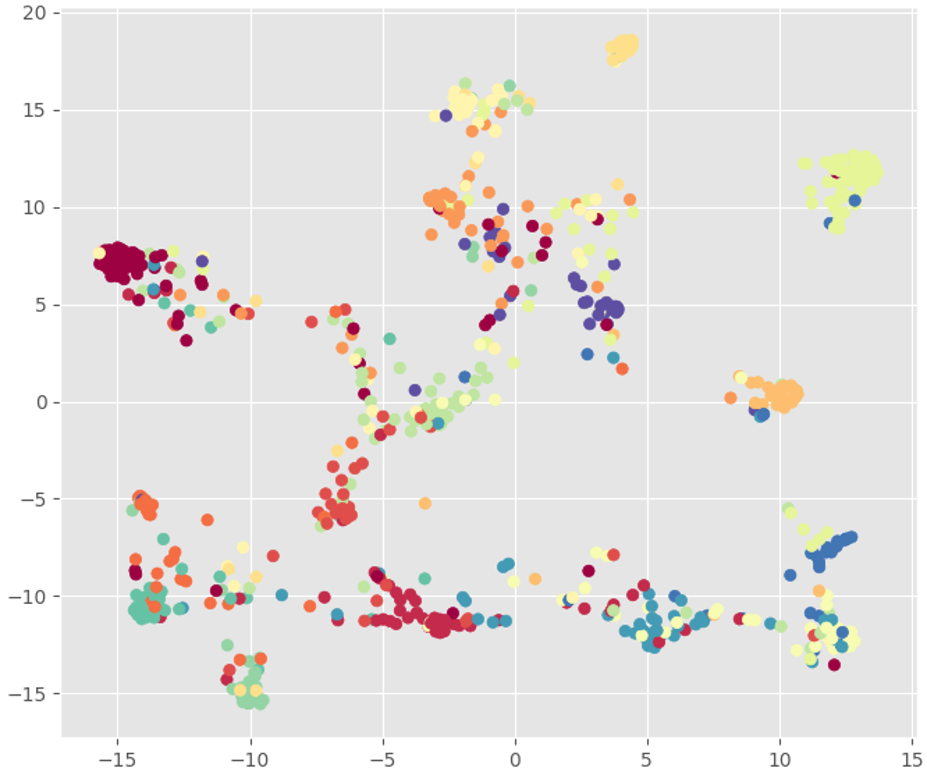}
   \caption{Singapore~\cite{style_classification}, Categorical.}
   \label{subfig_tsne_siam_hu}
   \end{subfigure} 
\caption{t-SNE~\cite{tsne} visualizations of the test sets of the Bonn Furniture Styles Dataset and Singapore Furniture Dataset~\cite{style_classification}.  Colors refer to the different styles.}
\label{fig_tsne}
\end{figure}

In order to visually inspect the quality of the learned mappings and compare the difficulty of the datasets, we visualize the test sets of the Bonn Furniture Styles Dataset and Singapore Furniture Dataset~\cite{style_classification} in a 2D space using dimensionality reduction (see Fig.~\ref{fig_tsne}).  We compare the categorical Siamese network and the fine-tuned base network GoogLeNet~\cite{googlenet}. For the categorical Siamese, we first map the images into the 256-dimensional embedding space, and then perform dimensionality reduction on the embedding vectors using t-distributed Stochastic Neighbour Embedding (t-SNE)~\cite{tsne}. For the Fine-tuned GoogLeNet~\cite{googlenet} we extract the features before the classification layer and apply t-SNE~\cite{tsne}. Fine-tuned GoogLeNet~\cite{googlenet} tends to produce several sub-clusters for the same style category~(Fig.~\ref{subfig_tsne})  while categorical Siamese tends to create single large clusters~(Fig.~\ref{subfig_tsne_siam}).  Compared the Singapore Furniture Dataset~\cite{style_classification} dataset, our dataset contains more images and is stylistically more diverse.

\subsection{Style-based comparison of all models}
Table~\ref{table_style_auc}   depicts the AUC scores of all Siamese networks on the compatibility task over the 17 style classes.
\begin{table}
\centering
\begin{tabular}{|l|c|c|c|}
\hline
  Style &
  \multicolumn{1}{l|}{\begin{tabular}[c]{@{}l@{}}Canonical\\ Siamese\end{tabular}} & \multicolumn{1}{l|}{\begin{tabular}[c]{@{}l@{}}Categorical\\ Siamese\end{tabular}} & \multicolumn{1}{l|}{\begin{tabular}[c]{@{}l@{}}Short\\ Siamese\end{tabular}} \\ 
  \hline
  Asian  & \textbf{0.84} & 0.83 & 0.75\\ \hline
  Beach  & 0.66 & \textbf{0.70} & 0.62\\ \hline
  Contemporary & \textbf{0.74} & \textbf{0.74} & 0.72 \\ \hline
  Craftsman & 0.76 & \textbf{0.77} & 0.76\\ \hline
  Eclectic & 0.59 & 0.61 & \textbf{0.62}\\ \hline
  Farmhouse & \textbf{0.71} & \textbf{0.71} & 0.65\\ \hline
  Industrial & \textbf{0.82} & \textbf{0.82} & 0.76\\ \hline
  Mediterranean & 0.73 & \textbf{0.76} & 0.66 \\ \hline
  Midcentury & 0.84 & \textbf{0.86} &  0.84  \\ \hline
  Modern  & \textbf{0.75} & \textbf{0.75} & 0.73 \\ \hline
  Rustic  & 0.62 & \textbf{0.64} & 0.63\\ \hline
  Scandinavian & 0.72 & 0.74 & \textbf{0.75} \\ \hline
  Southwestern & 0.59 & \textbf{0.62} & 0.56 \\ \hline
  Traditional & 0.72 & \textbf{0.73} & 0.69 \\ \hline
  Transitional & \textbf{0.75} & 0.73 & 0.71 \\ \hline
  Tropical & \textbf{0.63} & 0.61 & 0.51 \\ \hline
  Victorian & \textbf{0.87} & \textbf{0.87} & 0.83 \\ \hline
\end{tabular}
\caption{AUC scores of the Siamese networks on the compatibility task over 17 style classes in the Bonn Furniture Styles Dataset }
\label{table_style_auc}
\end{table}

\bibliographystyle{splncs04}
\bibliography{egbib}